\documentclass[5p]{elsarticle}

\usepackage{amssymb}
\usepackage{amsmath}

\journal{Neurocomputing}

\usepackage{xspace}
\makeatletter
\DeclareRobustCommand\onedot{\futurelet\@let@token\@onedot}
\def\@onedot{\ifx\@let@token.\else.\null\fi\xspace}

\def\eg{\emph{e.g}\onedot} 
\def\ie{\emph{i.e}\onedot} 
 
\def\etc{\emph{etc}\onedot}

\makeatother

\biboptions{sort&compress} 
\usepackage{graphicx}
\usepackage{caption}
\usepackage{float}
\usepackage{subcaption}
\usepackage{tabularx}
\usepackage{multirow}
\usepackage{makecell}
\usepackage{algorithm}
\usepackage{algpseudocode}
\usepackage{array}
\newcolumntype{?}{!{\vrule width 1pt}}

\usepackage[breaklinks,colorlinks]{hyperref}
\usepackage[capitalize]{cleveref}
\crefname{section}{Section}{Sections}
\crefname{table}{Table}{Tables}
\crefname{figure}{Figure}{Figures}

\newcommand{\mathbbm}[1]{\text{\usefont{U}{bbm}{m}{n}#1}}

\begin{document}
\begin{frontmatter}

\title{Calibrating Probabilistic Object Detectors with Annotator Disagreement} %

\author{Zhi Qin Tan\corref{cor1}}
\ead{zhi\_qin.tan@kcl.ac.uk}
\author{Owen Addison}
\author{Yunpeng Li}

\affiliation{organization={Faculty of Dentistry, Oral \& Craniofacial Sciences, King's College London},%
            city={London},
            country={United Kingdom}}
\cortext[cor1]{Corresponding author}

\begin{abstract}
High degrees of disagreement among annotators can exist for ambiguous objects, e.g. in medical images, underscoring the challenges of establishing ground truth annotations in object detection tasks. Despite this, all existing object detectors implicitly require access to ground truth annotations for either training or evaluation. 
The fundamental questions we target are: How can we learn an object detector with multiple annotators' annotations but without objective ground truth annotations due to object ambiguity, and how can we enable the learned detector to express meaningful model predictive uncertainties in detecting ambiguous objects? 
To answer these questions, we present an interpretable approach to calibrate probabilistic object detectors, where the calibration goal is to align the class confidence and bounding box variance estimates to the annotators' annotation distribution. 
We introduce an efficient yet effective framework to calibrate probabilistic object detectors by designing four evaluation metrics to measure calibration errors regarding classification and localization, and proposing a train-time calibration and post-hoc calibrator, all without the need to access any ground truth. This framework is generalizable to many existing probabilistic object detectors, such as the YOLO families and two-stage detectors. Empirical results with real-world and synthetic datasets of medical and natural images demonstrate the superior performance of the proposed framework with three popular object detectors. 
\end{abstract}


\end{frontmatter}

\section{Introduction}
\label{sec:intro}
\begin{figure}[tb]
    \centering
    \begin{subfigure}{.95\linewidth}
        \centering \includegraphics[width=\linewidth]{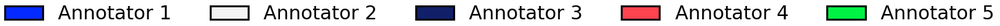}
    \end{subfigure}
     \begin{subfigure}{0.50\linewidth}
        \centering \includegraphics[width=\linewidth]{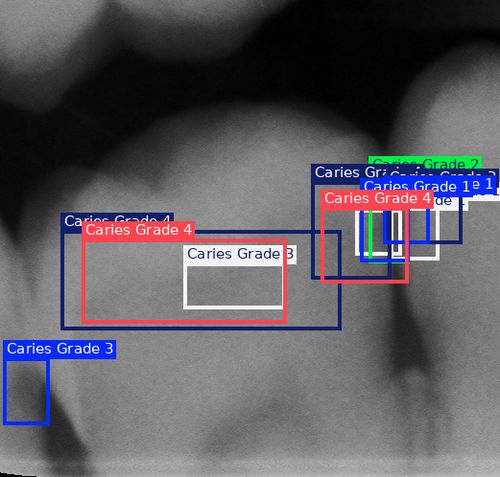}
    \end{subfigure}
    \hfill
     \begin{subfigure}{0.487\linewidth}
        \centering \includegraphics[width=\linewidth]{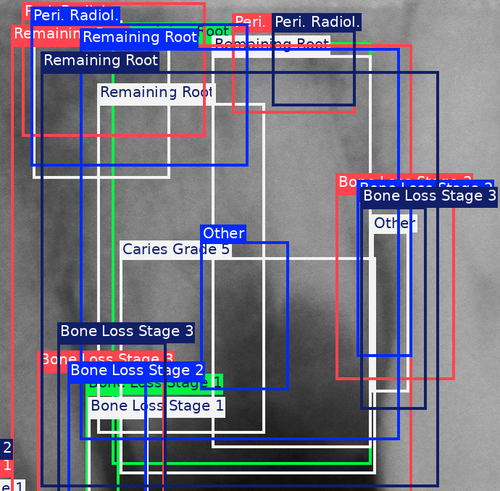}
    \end{subfigure}
    \begin{subfigure}{0.7\linewidth}
        \centering \includegraphics[width=\linewidth]{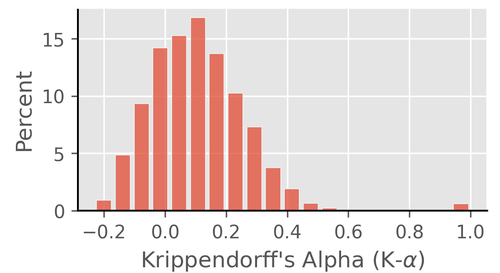}
    \end{subfigure}
    \caption{Top: Visualization of annotations on the DX-20 dataset. Annotations from different annotators are shown in different colors. 
    Bottom: Histogram of the inter-annotator agreement score, K-$\alpha$.
    }
    \label{fig:teaser_fig}
\end{figure}

Object detection is a common task in various applications, including medical imaging \cite{vindr2022,dentex2023,deeploc2024} and natural images \cite{pascal-voc-2007,cocodataset,waymo2020}. Conventional object detectors are trained with properly annotated ground truths, often acquired using majority voting of multiple annotators \cite{cocodataset}. Nevertheless, in complex domains such as medical images, interobserver variability \cite{interobsstudy2023} and disagreement among expert annotators lead to difficulty when defining ground truths. The disagreement can be due to differences in annotator expertise, viewing conditions, and more fundamentally, object ambiguity. For instance, a 2D X-ray image displays averaged projections of 3D objects, which often makes it impossible to annotate an early disease object with complete certainty. 
To illustrate this, we visualize in \Cref{fig:teaser_fig} real-world disease annotations of dental X-rays in our DX-20 dataset (See \Cref{sec:experiments} for details of the dataset). We observe a high degree of annotation disagreement on the same images by different expert dentists. This is evident through an average Krippendorff's Alpha (K-$\alpha$) score \cite{david_krippendorff2022} of 0.109, indicating extremely low inter-annotator agreement in the DX-20 dataset, which makes it highly challenging to obtain ground truth. 
In this paper, we define the ground truth as the single final true label, which is often expensive and difficult to obtain. For example, in the medical imaging domain, ground truth is usually established through a panel discussion to unify multiple expert annotations.
Several prior works \cite{crowdrcnn2020,Le2023,tan2024bdc} investigated improved annotation aggregation strategies using crowdsourced annotations. However, aggregated annotations to approximate ground truth may overlook valuable information revealed by disagreement between annotators, which can be useful for model training and calibration. %

One direction of capturing annotation uncertainty is akin to the task of calibrating object detectors. Conventional methods rely on or implicitly assuming access to ground truth annotations to evaluate calibration results, whereas our problem setup does not assume such access. Specifically, existing approaches for calibrating object detectors typically aim to match the model confidence to the precision of the detector (D-ECE \cite{d_ece2020}) or consider the joint task of matching the precision and intersection over union (IoU) of the detector (LaECE \cite{la_ece2023}). 
While these methods are widely used, they require ground truth annotations to compute the precision and IoU. Additionally, explaining concepts such as precision and IoU to a lay audience may not be straightforward. In contrast, annotators' annotation distribution offers a more intuitive approach to interpret and explain model confidence to end-users involved in subsequent decision-making \cite{Mehrtash2020qi,CHEN2022102444,Gawlikowski2023}. 

To the best of our knowledge, there was only one initial attempt on the calibration of probabilistic object detectors despite its significance, where \citet{dmm2019} first converted each element of the bounding box distribution to a confidence score using the inverse cumulative distribution function, then calibrated the predicted confidence to the empirical confidence similar to the calibration setup in classification tasks. However, the method requires the ground truth, and the predicted bounding box variances remain uncalibrated, as their method operates directly on the output of the inverse cumulative distribution function, overlooking the underlying bounding box distribution.

In this work, we offer a new perspective on \emph{calibrating object detectors by leveraging annotators' annotation uncertainty}. Specifically, we present a calibration framework for probabilistic object detectors where the objective is to align the predicted class probabilities and bounding box variances to the annotation distributions among the annotators, \emph{without using any ground truth annotations}. Probabilistic object detectors are used as they can output probability distributions for both object localization and object classification tasks, in contrast to deterministic object detectors that only output class probabilities and point estimates of bounding boxes. 
Our main contributions include:

\begin{itemize}
    \item We formally define the task of calibrating probabilistic object detectors with annotators' annotation uncertainty and propose four evaluation metrics to measure the alignment between the model's predictive uncertainty and the annotation distribution am\-ong the annotators, without the requirement to access ground truth annotations.
    \item We propose a simple but effective calibration method for probabilistic object detectors 
    by designing training loss functions that align with the calibration objective 
    and incorporating isotonic regression for post-hoc calibration.
    \item Extensive experiments demonstrate the effectiveness of the proposed method on both real and synthetic datasets of medical and natural images. Moreover, the method introduces negligible computational overhead and achieves an improvement in the mean calibration metric by $4.0 - 11.5$ in the experimental setups evaluated, while preserving detection accuracy.
\end{itemize}

\section{Related Work}
\label{sec:related}
\paragraph{\textbf{Object Detection with Noisy Annotations}}
Recent object detection research increasingly focuses on mitigating noisy annotations across the entire pipeline. To understand data limitations, \citet{nowak_cvprw2025} analyzed the inter- and intra-observer variability and proposed a validation method to identify noisy annotations from human annotators, while \citet{tschir_wacv2025} used contradictory annotations to define theoretical performance upper bounds for object detection datasets. Moreover, distribution-aware calibration \cite{Zhou_2024_BMVC} to handle bounding box inaccuracies for robust model training and noise-aware evaluation protocols \cite{murru_wacv2025} for reliable assessment under noisy ground truth are also introduced.

Besides, noise is also present when data is collected through crowdsourcing or compiled from multiple annotators. Several previous works \cite{crowdrcnn2020,Le2023,tan2024bdc} proposed to leverage crowdsourced object annotations \cite{crowdsourcing2012su} by aggregating them to approximate the ground truth. For example, \citet{crowdrcnn2020} proposed Crowd R-CNN, where the crowdsourced annotations are first pre-processed via a clustering algorithm, then learning the sensitivity of each annotator to different object classes. Besides, in \cite{tan2024bdc}, the authors model the annotators' annotations with categorical and Gaussian distributions, then aggregate the annotations based on the learned posterior distributions. These methods are not applicable to the proposed work as they are designed for deterministic object detectors and disregard the annotator disagreement information when annotations are aggregated.

\paragraph{\textbf{Calibration of Deterministic Object Detectors}}
Generally, calibration refers to the alignment of model accuracy and confidence for the classification task \cite{guo2017oncalib,kumar2019verif,nixon2019measur,mukh2020calib,wang2021rethink}. 
An object detector is considered calibrated if it has a precision of $p$ for all predictions with confidence of $p$, $\forall p\in [0,1]$ \cite{d_ece2020}.
LaECE \cite{la_ece2023} further enhanced the calibration objective by considering the product of model precision and average IoU. %
\citet{kuzucu2024calibration} proposed LaACE$_0$, utilizing an adaptive binning approach instead of a fixed number of bins.
Further, \citet{park2026uncertainty} introduced Object-level Calibration Error, which overcomes the weakness of prior calibration metrics on Detection Transformer (DETR) models that produce a much larger number of predictions than the number of ground truth objects.
Although these evaluation metrics are related, their calibration objectives are fundamentally different from ours, \ie they match model confidence to the model precision and IoU, and ignore the bounding box distribution, while our goal is to align model confidence with annotators' agreement. 

Prior methods for calibrating deterministic object detectors can be grouped into two types: 1) Train-time calibration techniques which typically introduce an auxiliary regularization loss to calibrate confidence scores during the training stage \cite{munir2022towards,pathi2023multiclass, munir2023caldetr, munir2023bridg,popor2024beyondcl}.;
2) Post-hoc calibration approaches which fit a calibration function that maps the predicted confidence to the calibrated confidence scores \cite{guo2017oncalib,la_ece2023}. Recent work \cite{kuzucu2024calibration} has also shown that when implemented correctly, post-hoc methods can yield better calibration results compared to train-time calibration methods. On the other hand, \citet{alexandridis2024fractal} proposed FRACAL, which aims to calibrate model outputs to improve object detection of rare categories. These methods are not directly applicable to our task as they all require ground truth annotations.

\paragraph{\textbf{Calibration of Probabilistic Object Detectors}}
Unlike deterministic object detectors, probabilistic object detectors aim to quantify uncertainty for object localization and classification \cite{poddef2020hall,probobjdetreview2022}. They can be broadly categorized into two groups: 1) Sampling-based methods \cite{dropoutsamp2018,evalmergestrat2019,probdeepensemble2020} which utilize Monte Carlo approximation of Bayesian neural networks \cite{bayesneurnetwork} (\eg ensemble of deep neural networks \cite{probdeepensemble2020}) to produce multiple object predictions, and approximate the predictive uncertainty;
2) Gaussian object detectors \cite{uncestdeepobjdet2018, gaussyolov32019,bboxregunc2019he,unconestage2019,bayesod2020,deepmixden2020,estandevaldod2021,uncquanselfdrive2023su} directly model the distribution of bounding boxes by treating them as Gaussian variables, thus enabling the models to capture bounding box uncertainty in object localization. 
Probabilistic object detectors are evaluated using the probability-based detection quality (PDQ) \cite{poddef2020hall} measure. 
\citet{dmm2019} present the first attempt to calibrate probabilistic object detectors by using the confidence computed with the inverse cumulative distribution function to estimate a confidence interval, then aligning the class confidence and the percentage of ground truth bounding boxes within the estimated interval. This method is not directly applicable to our work as it requires the ground truth position of object instances.

\section{Calibration of Probabilistic Object Detectors}
\label{sec:methodology}
In this section, we first introduce the calibration task of matching probabilistic object detection uncertainty outputs to the annotation distributions of the annotators. We then define four evaluation metrics suitable for the task, all without the presence of ground truths. Note that the calibration task addresses multiple expert annotations (\ie $>1$ annotations for each object instance) but does not require the final definitive ground truth (\ie one annotation only for each object instance).

\subsection{Problem Statement and Notations} 
\label{subsec:problem}
We assume a dataset $\{(x^i, y^i)\}_{i=1}^N$ with pairs of input image $x^i$ and annotations from multiple annotators $y^i = \{y^i_m\}_{m=1}^{M^i}$, where $N$ is the total number of training samples and $M^i$ is the total number of annotations of the $i$-th image from all annotators. 
Each annotation is a tuple of a bounding box, class label, and the annotator ID, $y^i_m = (b^i_m, c^i_m, k^i_m)$ in which $b^i_m \in \mathbb{R}^4$ contains the x-y coordinates of the top-left and bottom right corners of the box, $c^i_m \in \{1,...,J\}$ represents the annotated class label of the object and $k^i_m \in \{1,...,K\}$ is the annotator ID. $J$ denotes the number of object classes while $K$ is the total number of annotators. Note that each annotator can have annotation sets of varying cardinality. For brevity, notations in the rest of the paper are simplified to a single training sample. For the remainder of the paper, superscripts are used to index vectors or matrices.

The Gaussian object detector outputs a varying number of $P$ predictions, $\hat{y} = \{(\hat{\mu}_n, \hat{\Sigma}_n, \hat{p}_n)\}_{n=1}^{P}$, given an input image, $x$, where $\hat{\mu}_n \in \mathbb{R}^4$ is the predicted bounding box mean as the x-y coordinates of the top-left and bottom right corners, $\hat{\Sigma}_n \in \mathbb{R}^{4 \times 4}$ is the predicted bounding box covariance (a symmetric positive semi-definite matrix), and $\hat{p}_n \in [0, 1]^{J}$ is the predicted class probability of the $n$-th object. The bounding box and class label are modeled as Gaussian and categorical variables, respectively, in a Gaussian object detector. 
We aim to learn a Gaussian object detector to predict an uncertainty output (\ie class probability and bounding box distributions) that aligns with the annotators' annotation distributions.

\subsection{Pre-processing of Annotations}
\label{subsec:preprocess}
Before presenting the calibration metrics, we introduce a pre-processing module that forms groups of annotations representing annotations of the same object instance from different annotators. This module constructs sets of \emph{reference annotation} to train and evaluate a Gaussian object detector. 

Similar to previous approaches \cite{david_krippendorff2022,crowdrcnn2020,tan2024bdc}, we first employ a clustering algorithm to pre-process the annotations from multiple annotators. We utilize the Hungarian algorithm to perform annotation matching by assuming that the annotations from different annotators are unique sets of annotations. Then, to perform matching between any two pairs of annotation sets, we compute a cost matrix between the entries with the IoU cost function, $1 - IoU(b_i, b_j)$, where $b_i$ and $b_j$ are the bounding boxes of any annotation pair from the two annotator sets. While the optimal solution for a multipartite graph ($>2$ annotators) does not exist \cite{apx}, we can assume that for a sufficiently small number of annotators, an optimal solution can be found with a simple greedy matching \cite{david_krippendorff2022}.
After matching the annotations, we obtain annotation clusters, $A = \{a_h\}_{h=1}^H$, representing annotations from different annotators for the object instances in the image. Here, $H$ is the total number of clusters, and $a_h$ is a set containing tuples of bounding box, $b_h$, and class label, $c_h$, that belong to the $h$-th annotation cluster. 

Then, for each cluster of annotations, we compute its soft class probability label as follows:
\begin{equation}
    t_h^j = \frac{1}{K} \sum_{(b_h, c_h) \in a_h} \mathbbm{1}[c_h=j] \,\,,
    \label{eq:soft_cls_target}
\end{equation}
where $t_h^j$ is the target class probability of the $j$-th class for the $h$-th cluster, 
and $\mathbbm{1}[\cdot]$ is the indicator function. 
When a cluster does not contain an annotation from a certain annotator, they are assumed to have annotated the `background' class (with class index $j=0$) for the cluster. Hence, we obtain a class probability vector of length $J+1$, \ie $t_h \in [0, 1]^{J+1}$. 
Finally, we calculate the top-left and bottom-right corners' mean x-y coordinates of each annotation cluster, denoted as $\bar{b}_h$ for the $h$-th cluster, which is used to match each annotation cluster to the predicted bounding boxes for model training and evaluation. 

\subsection{Measuring Calibration Error}
\label{subsec:eval}
Given a test image with annotation clusters from the pre-processing module and object predictions (bounding box mean, covariance matrix, and class probabilities) from a Gaussian object detector, we propose four evaluation metrics to reward predictions that align well with the annotators' uncertainty. 
To this end, we first match the annotation clusters to the predicted bounding boxes by adopting the Hungarian algorithm. To account for the predicted bounding box distribution, we use the mean squared Mahalanobis distance \cite{mahalanobis1936generalized} of each annotated bounding box in the cluster, $a_h$, and the predicted distribution, represented as $\hat{\mu}_h$ and $\hat{\Sigma}_h$, as the cost function:
\begin{equation}
    \ell_h = \frac{1}{K} \sum_{(b_h, c_h) \in a_h} (b_h-\hat{\mu_h})^T\hat{\Sigma_h}^{-1}(b_h-\hat{\mu_h})  \ \ .
\end{equation}
Since the optimization algorithm will always match a cluster when there are more predicted objects, we modify the matching algorithm to avoid matches with a zero IoU between the annotated bounding box mean and the predicted bounding box mean.
Similar to the COCO evaluation metric, the matching procedure results in true positives (TP, a prediction-annotation cluster match), false positives (FP, a prediction with no matched annotation cluster), and false negatives (FN, an annotation cluster without a corresponding prediction).

\paragraph{\textbf{Classification metrics}} We evaluate the model's classification uncertainty by computing the total variation distance (TVD) \cite{tvd} between the predicted class probabilities and soft class probabilities labels of the matched annotation cluster:
\begin{equation}
    TVD = \frac{1}{|S|} \sum_{(t, \hat{p}) \in S}\ \frac{1}{2} ||t - \hat{p}||_1 \,\,,
    \label{eq:tvd}
\end{equation}
where $S$ is the set that contains all pairs of target and predicted class probability vectors.
TVD is chosen as the evaluation metric because it is easy to interpret and bounded within 0 and 1, though any distance or divergence metrics that quantify the statistical distance between two categorical distributions can be considered (\eg, Jensen–Shannon divergence). 
When no match is found for either the object prediction (\ie FP) or the annotation cluster (\ie FN), the missing component in \Cref{eq:tvd} is set to the one-hot encoding for the background class. Since the number of TP and FN is the same across all object detectors, we report their TVD metric together (\ie $S$ is the union of all TP and FN instances). For FP, we substitute $S$ with the set of all FP instances and compute the TVD\textsubscript{FP} metric separately, as different object detectors can predict varying numbers of objects on the same image, resulting in different amounts of FP. 
Note that \Cref{eq:tvd} includes the background class probability ($j=0$) to reward object detectors that correctly predict the probability of annotators not annotating the object instance. 

\paragraph{\textbf{Localization metrics}}

To evaluate the model's localization uncertainty, we compute the object detector's prediction certainty (\ie, confidence score of the predictions containing an object), $\hat{o}$, either by using the predicted foreground probability (one minus the background probability), the objectness score or the maximum predicted class probability, depending on the model architecture. We note that as different architectures have different semantic meanings of prediction certainty, the calibration metrics should only be used to compare against their uncalibrated counterparts of the same architecture. The prediction certainty is used as the confidence interval along with the predicted bounding box mean and variance to compute the predicted bounding box interval, $[\hat{l}, \hat{u}]$ \cite{conf_interval}. The localization uncertainty error (LUE) is then computed as the absolute difference between the prediction certainty and the percentage of annotated bounding boxes that is within the predicted bounding box interval:
\begin{multline}
    LUE = \frac{1}{|S_{TP}|} \sum_{(\hat{o}, \hat{l}, \hat{u}, a) \in S_{TP}} \Big|\hat{o} \\- \frac{1}{|a|}\sum_{(b, c) \in a} \mathbbm{1}[(\hat{l} < b) \land (b < \hat{u})]\Big| \,\,,
    \label{eq:lue}
\end{multline}
where $S_{TP}$ is the set of TP instances (prediction-annotation matches) containing the prediction certainty, $\hat{o}$, the predicted bounding box lower interval, $\hat{l} \in \mathbb{R}^4$, and upper interval, $\hat{u} \in \mathbb{R}^4$, that are matched to the annotation cluster, $a$, which contains tuples of annotated bounding box, $b$, and class label, $c$, while $\land$ is the ``AND'' boolean operator. 

Furthermore, to account for FN of the object detector, we compute a penalizing term denoted as FNE for the set of FN, $S_{FN}$, containing all annotation clusters without a prediction match as follows:
\begin{equation}
    FNE = \frac{1}{|S_{FN}|} \sum_{a \in S_{FN}} \frac{|a|}{K} \,\,.
    \label{eq:fne}
\end{equation}
Here, $h$ is a cluster index with no matched prediction.

\paragraph{\textbf{Overall}} The evaluation metrics are reported as TVD, TVD\textsubscript{FP}, LUE, and FNE, as well as their geometric mean. Each metric has a range between 0 and 1, with a score of 0 indicating a perfect capture of annotators' uncertainty, while a score of 1 indicates the opposite. Since lower is better for the evaluation metrics, we calculate the geometric mean of the metrics' complement (\ie, $1-metric$), then invert the result back to the original scale. This ensures that a model must perform well across all four metrics to achieve a low overall calibration error, and it heavily penalizes the model if any single component performs poorly (approaching one). We note that, depending on the application or imaging domain, different weights can be applied to the four metrics to balance the importance of calibrating the classification versus localization uncertainty.

\section{Calibration Methodology}
\label{sec:calib_method}

\begin{figure*}[tb]
    \centering
    \includegraphics[width=\linewidth]{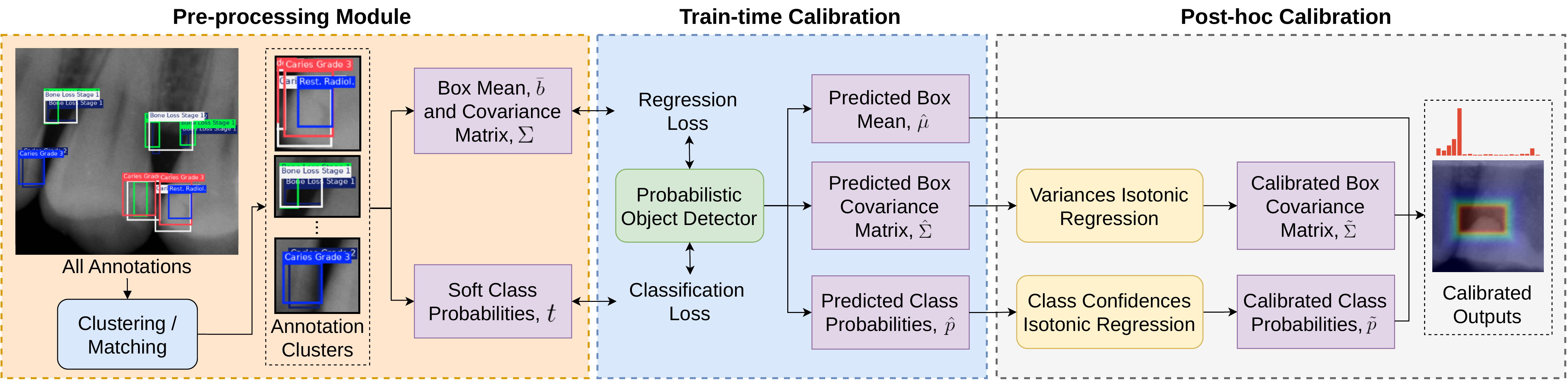}
    \caption{Overall framework of the proposed calibration method, where all annotations are first pre-processed to obtain annotation clusters, then a combination of train-time and post-hoc calibration is employed to produce the final calibrated predictions. Different annotators are visualized in different colors in the pre-processing module. Best viewed in color.}
    \label{fig:overall_method}
\end{figure*}

\Cref{fig:overall_method} presents an overview of our proposed calibration methodology, which combines train-time and post-hoc calibration to align the model's uncertainty output with the annotators' uncertainty. 
We first train a Gaussian object detector by learning the categorical and Gaussian distributions of the annotations clusters in a supervised fashion, then adopt post-hoc isotonic regression to calibrate the predicted distributions further to match the annotators' agreement.

\subsection{Train-time Calibration}
\label{subsec:traintime}

Before we can compute the training loss terms, the matching algorithm that the base Gaussian object detector adopts is applied between the predicted bounding box mean, $\hat{\mu}_n$, and the bounding box mean of each annotation cluster, $\bar{b}_h$, from the pre-processing module (\Cref{subsec:preprocess}). As a result, we obtain pairs of object predictions and annotation clusters to compute classification and localization regression losses.

\paragraph{\textbf{Classification loss}} Following previous Gaussian object detectors, the loss function used in their deterministic counterpart, \ie cross-entropy loss, is used to train the classification head (and objectness head for YOLO models) \cite{uncestdeepobjdet2018, gaussyolov32019,bboxregunc2019he,unconestage2019,bayesod2020,deepmixden2020,estandevaldod2021}:
\begin{equation}
    \mathcal{L}_{c}(t_h, \hat{p}_h) = - \sum_{j=1}^J t_h^j \ln \hat{p}_h^j \,\,,
    \label{eq:class_loss}
\end{equation}
where $\hat{p}_h$ is the predicted class probability that is matched to the $h$-th annotations cluster. 
The key difference is that our training target is the soft class probability labels, $t_h$, of the $h$-th annotation cluster instead of hard one-hot labels. 
Depending on the model architecture, the additional $+1$ background class probability (\ie $t_h^0$) can be left as is (\eg Faster R-CNN \cite{fasterrcnn2015}), converted to objectness score target as one minus the background probability (\eg YOLO models \cite{yolox2021, wang2022yolov7}) or removed entirely (\eg RetinaNet \cite{retinanet2017}).

\paragraph{\textbf{Regression loss}} For the object localization regression task, existing Gaussian object detectors \cite{uncestdeepobjdet2018, gaussyolov32019,bboxregunc2019he,unconestage2019,bayesod2020,deepmixden2020,estandevaldod2021} 
utilize the negative log-likelihood (NLL) loss to learn the mean and covariance matrix of the predicted Gaussian distributions by treating the ground truth bounding boxes as individual samples from the Gaussian distributions. However, this neither captures the annotators' uncertainty nor aligns the annotators' bounding box distributions with the predicted distributions. 
Moreover, LUE (\Cref{eq:lue}) is not differentiable and cannot be used directly as the loss function. 
Instead, we propose a modification to the Direct Moment Matching (DMM) \cite{estandevaldod2021} loss function to align the annotators' bounding box distributions with the predicted distributions. Specifically, we match the predictive distribution to the annotators' statistics, rather than the sample statistics obtained using the ground truth values. For each annotation cluster, we first compute the target variance, $\sigma_h$, for the $h$-th cluster as follows:

\begin{align}
    \alpha_h = \min\left\{\frac{|a_h|}{K}, \gamma\right\}& \,,\quad  z_h = F^{-1}\left(1 - \frac{1-\alpha_h}{2}\right)
    \label{eq:alpha}
\end{align}
\begin{align}
    \sigma_h =&\ \left(\frac{\grave{b_h} - \acute{b_h}}{2 z_h}\right)^2 \,\,.
    \label{eq:box_var}
\end{align}
Here, $|a_h|/K$ is the annotator certainty (\ie percentage of annotators agreeing on the bounding box location in the $h$-th cluster) and $\gamma$ is a maximum confidence threshold set at $0.999$ for numerical stability. $F^{-1}(\cdot)$ is the inverse cumulative distribution function of a standard Gaussian distribution while $\acute{b_h} \in \mathbb{R}^4$ and $\grave{b_h} \in \mathbb{R}^4$ are the minimum and maximum bounding box values of the $h$-th cluster, $a_h$, respectively. 

In short, \Cref{eq:alpha,eq:box_var} computes the expected variance given the confidence interval and the specified level of certainty (annotator certainty).
The variances of each bounding box component are calculated to construct the target diagonal covariance matrix, $\Sigma_h$. 
Then, the localization loss is defined as:
\begin{equation}
    \mathcal{L}_{r}(\bar{b}_h, \Sigma_h, \hat{\mu}_h, \hat{\Sigma}_h) = ||\bar{b}_h - \hat{\mu}_h||_1 + ||\Sigma_h - \hat{\Sigma}_h||_1 \,\,,
    \label{eq:reg_loss}
\end{equation}
where $\hat{\mu}_h$ and $\hat{\Sigma}_h$ are the predicted bounding box mean and covariance matrix matched to the $h$-th annotations cluster, respectively, while $||\cdot||_1$ is the L1 norm. In short, given the annotator certainty, we train the model to predict a mean and covariance matrix such that all bounding boxes of the $h$-th cluster are within the predicted $\alpha_h$ confidence interval.

Finally, the parameter of the Gaussian object detector is optimized by minimizing the loss terms of each pair of object prediction and annotation cluster:
\begin{equation}
    \mathcal{L}_{t} = \lambda \mathcal{L}_{r} + \mathcal{L}_{c} \,\,,
    \label{eq:overall_loss}
\end{equation}
where $\lambda$ is a hyperparameter to balance the two loss terms. \Cref{alg:training} provides the pseudocode of the training steps for a single image.

\begin{algorithm}
    \small
    \caption{Pseudocode for the training steps of the train-time calibration method.}
    \label{alg:training}
    \begin{algorithmic}[1]
        \Require{The annotation clusters, $A=\{a_h\}^H_{h=1}$, where $H$ is the number of clusters; The object predictions, $\hat{y}=\{(\hat{\mu}_n, \hat{\Sigma}_n, \hat{p}_n)\}^P_{n=1}$, where $P$ is the number of predictions; The number of annotators, $K$; The weight loss balancing hyperparameter, $\lambda$.}
            \State $M \gets match(A, \hat{y})$ \Comment{Match the annotation clusters and object predictions with the base detector's matching algorithm.}
            \State $\mathcal{L} \gets 0$;
            \For {$i=1$ to $|M|$}
                \State $(a_i, \hat{y_i}) \gets M_i$;
                \State $(\hat{\mu}_i, \hat{\Sigma}_i, \hat{p}_i) \gets \hat{y}_i$;
                \State Compute the soft class probability, $t_i$, with \Cref{eq:soft_cls_target};
                \State Compute classification loss, $\mathcal{L}_c(t_i, \hat{p}_i)$, with \Cref{eq:class_loss};
                \State $\bar{b}_i = \frac{1}{|a_i|}\sum_{(b, c) \in a_i}b$;
                \State Construct the expected annotator covariance, $\Sigma_i$, according to \Cref{eq:alpha,eq:box_var};
                \State Compute regression loss, $\mathcal{L}_r(\bar{b}_i, \Sigma_i, \hat{\mu}_i, \hat{\Sigma}_i)$, with \Cref{eq:reg_loss};
                \State $\mathcal{L}_{t} = \lambda \mathcal{L}_{r} + \mathcal{L}_{c}$;
                \State $\mathcal{L} := \mathcal{L} + \mathcal{L}_t$;
            \EndFor
            \State $\mathcal{L} := \frac{\mathcal{L}}{|M|}$;
            \State Optimize model by minimizing the overall loss, $\mathcal{L}$;
    \end{algorithmic}
\end{algorithm}

\subsection{Post-Hoc Calibration}
\label{subsec:posthoc}
The predicted class confidences and bounding box variances of the trained Gaussian object detector are further calibrated with a post-hoc calibrator. We adopt isotonic regression (IR) \cite{robert1988isotonicreg} since it models the calibration task as a monotonic regression problem and is easy to implement with scikit-learn \cite{scikitlearn}.

\paragraph{\textbf{Training phase}} Using a trained Gaussian object detector, we obtain object predictions on a held-out validation set. Following the matching steps in \Cref{subsec:eval}, we can construct pairs of object prediction and annotation cluster to form the training dataset for the IR model. To prevent low-scoring predictions from dominating the training of the calibrator, we propose using the prediction certainty, $\hat{o}$, as the sample weight for each prediction during the training phase. The true annotators' agreement in terms of class probabilities and bounding box variances are calculated using \Cref{eq:soft_cls_target} and \Cref{eq:box_var}, respectively.

We train a class-specific confidence score calibrator $\zeta^j: [0, 1] \rightarrow [0, 1]$ for each class $j$ where the input is the predicted $j$-th class probability, $\hat{p}^j_n$, while the target is the percentage of annotators that annotated the $j$-th class, $t^j_n$. On the other hand, we train multiple variance calibrators $\xi^i: \mathbb{R^+} \rightarrow \mathbb{R^+}$ for each bounding box component, $i \in \{1, ..., 4\}$, using the predicted variance, $\hat{\Sigma}_n^{(i,i)}$, as input and the annotators' bounding box variance, $\sigma_n^i$, as the target. 
Overall, we obtain $J+4$ calibrators after the training phase.

\paragraph{\textbf{Inferencing phase}} Here, we consider a single instance of predicted class probability, $\hat{p}$, and bounding box covariance matrix, $\hat{\Sigma}$. When calibrating the class confidence during the inference stage, we must ensure that the calibrated class probability vector, $\Tilde{p}$, is still a valid categorical distribution, \ie the probability vector should sum to one after any calibration. To achieve this, for each calibrated confidence with predicted class $j$, we compute $\Tilde{p}^j = \zeta^j(\hat{p}^j)$ and modify the class probability of the other classes as follows:
\begin{equation}
    \Tilde{p}^c = \frac{\hat{p}^c}{(1 - \Tilde{p}^j)}\,, \forall c \in \{1, ..., J\} \land c \neq j \,\,.
    \label{eq:posthoc_cls_mod}
\end{equation}

Then, the calibrated bounding box covariance matrix, $\Tilde{\Sigma}$, is obtained by first setting $\Tilde{\Sigma}:= \hat{\Sigma}$, then applying the learned variance IR models:
\begin{align}
    \Tilde{\Sigma}^{(i,i)} =&\ \xi^i(\hat{\Sigma}^{(i,i)}), \forall i \in \{1, ..., 4\} \,\,.
    \label{posthoc_box_var}
\end{align}

\section{Experiments and Results}
\label{sec:experiments}

\subsection{Experiments Setup}
\paragraph{\textbf{Training details}}
To show that the proposed calibration framework can be generalized to different object detectors, we first extend YOLOX \cite{yolox2021} into a Gaussian object detector and use the probabilistic extension of Faster R-CNN \cite{fasterrcnn2015} and RetinaNet \cite{retinanet2017} implemented in \cite{estandevaldod2021,kendall_prob2017}.
We chose these 3 architectures to demonstrate the generalizability of our proposed framework to any object detector, including object detectors that predict objectness scores (YOLO detectors), a background class with the foreground classes (R-CNN detectors), and only the foreground classes (RetinaNet).
We then integrate them with the proposed framework, denoted as, YOLOX-G$\dag$, FRCNN-G$\dag$, and RetinaNet-G$\dag$, by following the training settings of the original YOLOX, Faster R-CNN, and RetinaNet. 

\Cref{table:hyperparams} shows the training settings for each object detection algorithm. During training, we apply random data augmentation such as rotation, translation, zooming, and horizontal flipping to the training images and annotations. The loss weight to balance the regression and classification losses, $\lambda$, is set to 0.1, similar to previous works in object detection (\eg YOLOX \cite{yolox2021}). We follow the training hyperparameters of the original implementation closely, but the batch size and learning rate are scaled according to the available GPU memory. 
Code \footnote{\url{https://github.com/zhiqin1998/calib-prob-od}} is implemented with PyTorch version 1.9 \cite{pytorch} and runs on an RTX 4090 GPU. 

\begin{table*}[tb]
    \centering
    \caption{Training configuration for Faster R-CNN (FRCNN) \cite{fasterrcnn2015}, RetinaNet \cite{retinanet2017}, and YOLOX \cite{yolox2021} object detectors (including both deterministic and probabilistic variants).}
    \small
    \resizebox{0.7\linewidth}{!}{
    \begin{tabular}{l|ccc}
        \hline
        Setting & FRCNN & RetinaNet & YOLOX \\
        \hline
        \hline
        Pre-training data & ImageNet & ImageNet & ImageNet  \\
        Learning rate (LR) & 0.001 & 0.001 & 0.0001 \\
        LR scheduler & Multi-step w/ warmup & Multi-step w/ warmup & Cosine decay w/ warmup \\
        Optimizer & AdamW & AdamW & SGD w/ momentum \\
        Weight decay & 0.0001 & 0.0001 & 0.0005\\
        Image size & 800 & 800 & 640 \\
        Batch size & 8 & 16 & 16\\
        Training iteration & 70k & 40k & 100k\\
        \hline
    \end{tabular}
    }
    \label{table:hyperparams}
\end{table*}

\paragraph{\textbf{Datasets}}
We evaluate detectors' performance in three different tasks and datasets, all with low annotator agreement.
Our in-house dataset, DX-20, consists of 3,755 dental radiographs from various dental practices with 20 classes representing different types of dental diseases, such as five grades of caries, four stages of bone loss, \etc. DX-20 is annotated by 28 dentists of various backgrounds and levels of experience. We randomly assigned five dentists to annotate each image. 
600 images with equal proportions from different dental practices were selected as the test dataset.

Besides, we evaluate on VBD-CXR \cite{vindr2022}, a thoracic abnormalities detection dataset consisting of 15,000 chest radiographs. VBD-CXR is annotated with 14 abnormality classes by 17 experienced radiologists, where three annotators are randomly assigned to each image. As the VBD-CXR test dataset is private, we perform a 9:1 split on the 15,000 images to obtain our test dataset. 

Moreover, we use VOC-MIX \cite{tan2024bdc}, a Pascal VOC \cite{pascal-voc-2007} dataset with synthetic crowdsourced annotations. In the VOC-MIX dataset, all 5,011 natural images were annotated by 25 simulated annotators of different annotating accuracy with 20 object classes (\eg car, chair, dog, person, \etc).%

\begin{table*}[tb]
    \centering%
    \caption{Calibration evaluation metrics for quantifying annotators' uncertainty. *-G and *-G$\dag$ denote Gaussian object detectors trained with the NLL loss and our proposed framework, respectively. Lower is better for all metrics, and bold indicates the best results.}
    \resizebox{\linewidth}{!}{
    \begin{tabular}{l?cccc|c?cccc|c?cccc|c}
        \hline
         \multirow{2}{*}{\textbf{Model}} & \multicolumn{5}{c?}{\textbf{DX-20}} & \multicolumn{5}{c?}{\textbf{VBD-CXR}} & \multicolumn{5}{c}{\textbf{VOC-MIX}} \\
          & TVD & TVD\textsubscript{FP} & LUE & FNE & Mean & TVD & TVD\textsubscript{FP} & LUE & FNE & Mean & TVD & TVD\textsubscript{FP} & LUE & FNE & Mean \\
         \hline \hline
         FRCNN & 37.4 & 32.0 & - & 36.1 & - & \textbf{38.0} & 40.1 & - & 45.5 & - & 41.8 & 38.4 & - & 56.9 & - \\
         FRCNN-DE & 39.7 & 31.5 & \textbf{40.2} & 35.5 & 36.8 & 39.7 & 29.2 & \textbf{18.7} & 45.8 & \textbf{34.1} & 40.3 & \textbf{36.5} & \textbf{23.9} & \textbf{43.7} & \textbf{36.5} \\
         FRCNN-G & 38.1 & 32.4 & 55.2 & 36.3 & 41.2 & 39.0 & 41.0 & 22.6 & 46.0 & 37.7 & 49.6 & 39.7 & 26.0 & 48.0 & 41.5 \\
         FRCNN-G$\dag$ (ours) & \textbf{36.6} & \textbf{29.0} & 41.3 & \textbf{31.7} & \textbf{34.8} & 44.9 & \textbf{27.3} & 19.0 & \textbf{44.8} & 35.0 & \textbf{38.6} & 38.4 & 24.6 & 44.7 & 37.0 \\
         \hline
         RetinaNet & 43.8 & 42.7 & - & 26.6 & - & 48.1 & 34.1 & - & 42.2 & - & 66.2 & 39.3 & - & 40.0 & - \\
         RetinaNet-DE & 41.2 & 32.5 & \textbf{43.2} & 25.9 & 38.6 & 47.3 & 33.8 & \textbf{26.0} & 42.3 & 37.9 & \textbf{60.4} & 37.0 & \textbf{49.4} & \textbf{36.3} & \textbf{46.8} \\
         RetinaNet-G  & 44.4 & 45.4 & 51.7 & \textbf{25.8} & 42.3 & 48.5 & 34.5 & 27.5 & 49.3 & 40.7 & 64.1 & 40.0 & 56.7 & 38.1 & 50.9 \\
         RetinaNet-G$\dag$ (ours) & \textbf{40.4} & \textbf{29.3} & 44.7 & 25.9 & \textbf{35.5} & \textbf{46.0} & \textbf{33.6} & 28.7 & \textbf{40.1} & \textbf{37.4} & 64.6 & \textbf{33.1} & 50.8 & 37.3 & 48.0 \\
         \hline
         YOLOX & 40.6 & \textbf{8.4} & - & 24.3 & - & 43.2 & 9.3 & - & 41.3 & - & 57.0 & 10.6 & - & 42.4 & - \\
         YOLOX-DE & \textbf{37.5} & 8.5 & \textbf{47.0} & 24.3 & \textbf{30.8} & \textbf{40.4} & 8.8 & \textbf{58.9} & 41.6 & \textbf{39.9} & \textbf{47.8} & \textbf{10.2} & 50.5 & 43.2 & \textbf{40.0} \\
         YOLOX-G & 40.6 & 8.9 & 51.4 & 24.0 & 33.1 & 43.5 & 9.5 & 65.6 & 41.5 & 43.4 & 53.0 & 10.4 & 56.5 & 45.3 & 43.8 \\
         YOLOX-G$\dag$ (ours) & 38.1 & 9.1 & 47.5 & \textbf{23.6} & 31.1 & 47.2 & \textbf{7.7} & 59.6 & \textbf{41.0} & 41.6 & 55.2 & 10.3 & \textbf{50.4} & \textbf{39.9} & 41.2 \\
         \hline
    \end{tabular}
    }
    \label{table:quantitative}
\end{table*}

Lastly, to understand the inter-annotator agreement of the datasets, we compute the K-$\alpha$ score \cite{david_krippendorff2022}. The K-$\alpha$ score evaluates the class level agreement after performing annotations matching between annotators, where 1 is a perfect agreement, and 0 means no agreement. DX-20, VBD-CXR, and VOC-MIX have a K-$\alpha$ of 0.109, 0.305, and 0.339, respectively. 
See \ref{appx:dataset} for further details of the datasets.

\paragraph{\textbf{Baselines}}
We compare the proposed method against multiple baselines. Since the training datasets only have annotations from multiple annotators and do not have real ground truths, we slightly modified them so that conventional object detectors can be used. First, we duplicate each training image and separate the annotations of different annotators. The duplicated images and separated annotations are then treated as individual data samples to train probabilistic object detectors with the NLL loss following \cite{estandevaldod2021}, denoted as FRCNN-G, RetinaNet-G, and YOLOX-G. 
Further, using the same duplicated dataset, we train the deterministic object detectors to compare them against their probabilistic counterparts, denoted by FRCNN, RetinaNet, and YOLOX.
Lastly, we adopt a deep ensemble \cite{deep_ensemble} of the deterministic object detectors, trained using labels from the different annotators of each dataset (\ie, five for DX-20, three for VBD-CXR, and five subsets of five annotators for VOC-MIX), denoted as FRCNN-DE, RetinaNet-DE, and YOLOX-DE. The object predictions from the different models are ensembled using the clustering procedure defined in \Cref{subsec:preprocess} to compute the predictive distribution of the class label and bounding box.

\subsection{Results and Discussion}
\label{sec:results}

\begin{table*}[tb]
    \caption{Comparison of our proposed method to calibration methods that are designed for deterministic object detectors, and with the presence of ground truth, evaluated with calibration metrics that require ground truth.
    For the DX-20 and VBD-CXR datasets, ground truth is approximated by majority voting, while for VOC-MIX, we use the original ground truth of the Pascal VOC dataset. 
    Lower is better for all metrics, and bold indicates the best results.}
    \label{table:stand_calib}
    \centering\setlength{\tabcolsep}{5pt}
    \small
    \resizebox{0.6\linewidth}{!}{
    \begin{tabular}{l|cc|cc|cc}
        \hline
        \multirow{2}{*}{\textbf{Model}} & \multicolumn{2}{c|}{\textbf{DX-20}} & \multicolumn{2}{c|}{\textbf{VBD-CXR}} & \multicolumn{2}{c}{\textbf{VOC-MIX}} \\
          & LaECE$_0$ & LaACE$_0$ & LaECE$_0$ & LaACE$_0$ & LaECE$_0$ & LaACE$_0$ \\
         \hline
         FRCNN & 20.7 & 30.0 & 39.9 & 44.5 & 18.0 & 27.5\\
         FRCNN-IR & \textbf{11.5} & \textbf{27.0} & \textbf{20.6} & 34.3 & \textbf{10.2} & \textbf{20.8} \\
         FRCNN-PS & 14.2 & 27.3 & 24.0 & \textbf{33.9} & 11.5 & 20.9 \\
         FRCNN-DE & 16.3 & 27.7 & 34.1 & 39.2 & 15.7 & 23.3 \\
         FRCNN-G$\dag$ (ours) & 16.7 & 27.6 & 31.3 & 37.2 & 14.6 & 22.5 \\
         \hline
         RetinaNet & 26.8 & 40.7 & 29.1 & 42.1 & 17.5 & 29.3 \\
         RetinaNet-IR & \textbf{14.7} & \textbf{27.1} & \textbf{18.3} & \textbf{30.0} & \textbf{10.7} & \textbf{20.8} \\
         RetinaNet-PS & 15.9 & 27.4 & 19.0 & 30.5 & 12.3 & 21.4 \\
         RetinaNet-DE & 15.5 & 28.9 & 21.3 & 32.1 & 14.9 & 24.6 \\
         RetinaNet-G$\dag$ (ours) & 15.3 & 29.7 & 22.5 & 32.6 & 15.2 & 23.1 \\
         \hline
         YOLOX & 18.5 & 27.8 & 20.1 & 35.7 & 17.8 & 28.8 \\
         YOLOX-IR & 14.4 & 19.1 & \textbf{15.3} & \textbf{27.4} & \textbf{9.0} & \textbf{20.6} \\
         YOLOX-PS & \textbf{13.8} & \textbf{18.8} & 15.8 & 28.5 & 10.8 & \textbf{20.6} \\
         YOLOX-DE & 14.5 & 20.8 & 17.8 & 29.3 & 13.4 & 22.7 \\
         YOLOX-G$\dag$ (ours) & 14.5 & 21.2 & 17.9 & 29.5 & 14.9 & 23.1 \\
         \hline
    \end{tabular}
    }
\end{table*}

\begin{figure*}[tbp]

    \begin{subfigure}{0.22\textwidth}
        \centering \includegraphics[width=\linewidth,trim={0cm 1cm 2cm 0},clip]{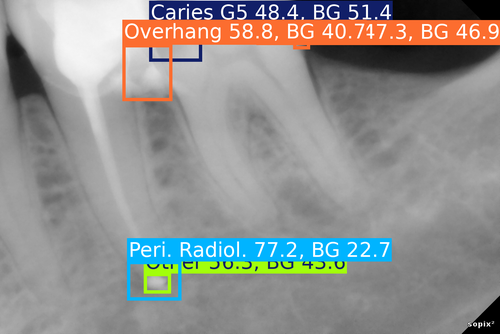}
    \end{subfigure} \hfill
    \begin{subfigure}{0.22\textwidth}
        \centering \includegraphics[width=\linewidth,trim={0cm 1cm 2cm 0},clip]{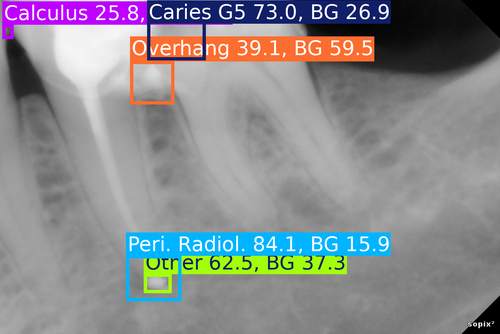}
    \end{subfigure} \hfill
    \begin{subfigure}{0.22\textwidth}
        \centering \includegraphics[width=\linewidth,trim={0cm 1cm 2cm 0},clip]{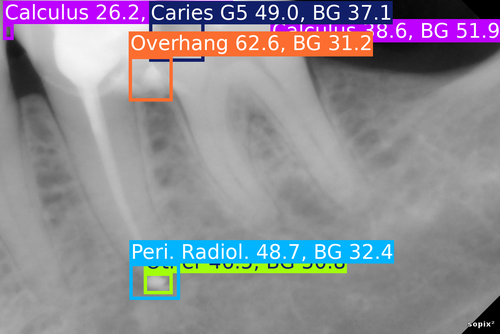}
    \end{subfigure} \hfill
    \begin{subfigure}{0.22\textwidth}
        \centering \includegraphics[width=\linewidth,trim={0cm 1cm 2cm 0},clip]{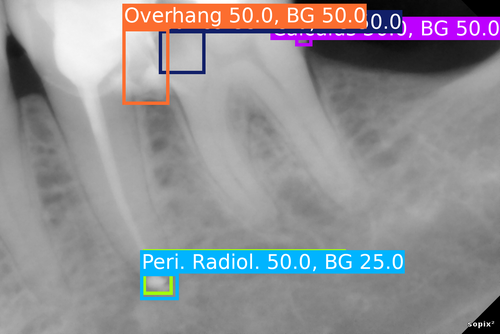}
    \end{subfigure}

    \smallskip
    \begin{subfigure}{0.22\textwidth}
        \centering \includegraphics[width=\linewidth,trim={3cm 2cm 0 7cm}, clip]{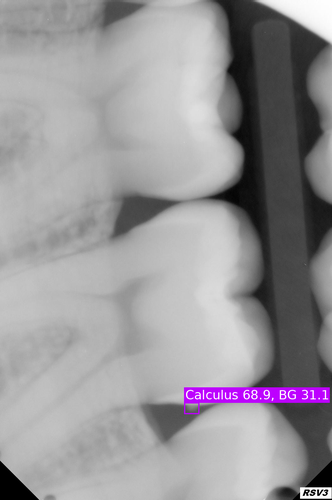}
    \end{subfigure} \hfill
    \begin{subfigure}{0.22\textwidth}
        \centering \includegraphics[width=\linewidth,trim={3cm 2cm 0 7cm}, clip]{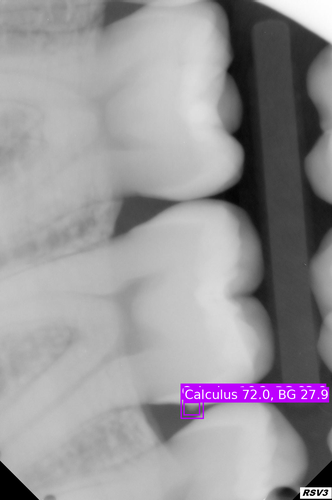}
    \end{subfigure} \hfill
    \begin{subfigure}{0.22\textwidth}
        \centering \includegraphics[width=\linewidth,trim={3cm 2cm 0 7cm}, clip]{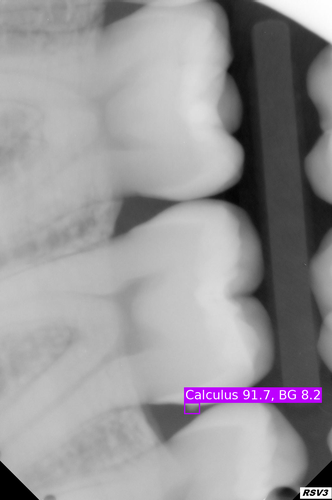}
    \end{subfigure} \hfill
    \begin{subfigure}{0.22\textwidth}
        \centering \includegraphics[width=\linewidth,trim={3cm 2cm 0 7cm}, clip]{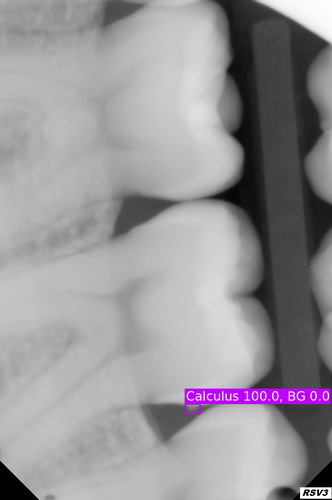}
    \end{subfigure}
    
    \smallskip
    \begin{subfigure}{0.22\textwidth}
        \centering \includegraphics[width=\linewidth,trim={0 2cm 0 1cm},clip]{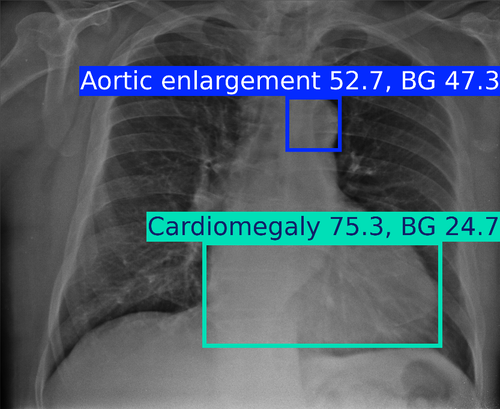}
    \end{subfigure} \hfill
    \begin{subfigure}{0.22\textwidth}
        \centering \includegraphics[width=\linewidth,trim={0 2cm 0 1cm},clip]{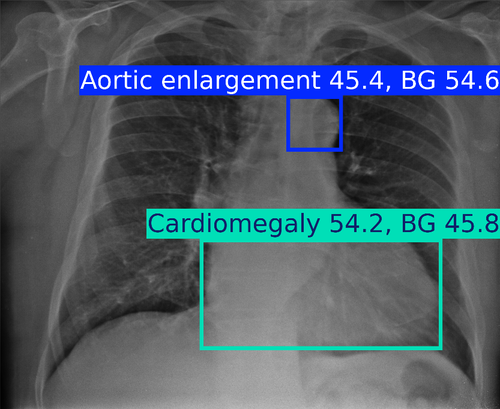}
    \end{subfigure} \hfill
    \begin{subfigure}{0.22\textwidth}
        \centering \includegraphics[width=\linewidth,trim={0 2cm 0 1cm},clip]{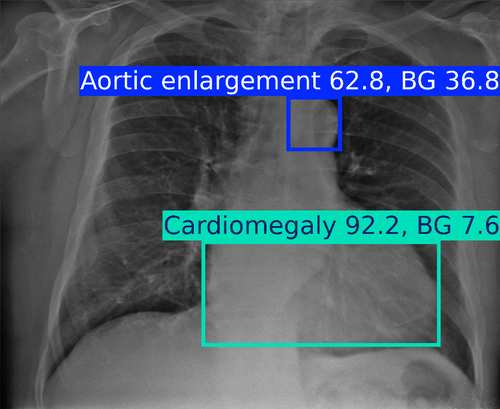}
    \end{subfigure} \hfill
    \begin{subfigure}{0.22\textwidth}
        \centering \includegraphics[width=\linewidth,trim={0 2cm 0 1cm},clip]{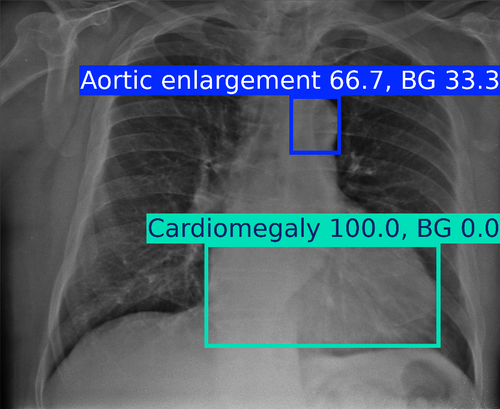}
    \end{subfigure}

    \smallskip
    \begin{subfigure}{0.22\textwidth}
        \centering \includegraphics[width=\linewidth,trim={0 2cm 0 2cm}, clip]{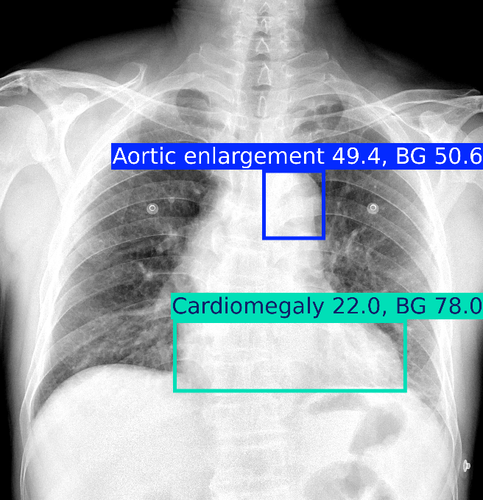}
    \end{subfigure} \hfill
    \begin{subfigure}{0.22\textwidth}
        \centering \includegraphics[width=\linewidth,trim={0 2cm 0 2cm}, clip]{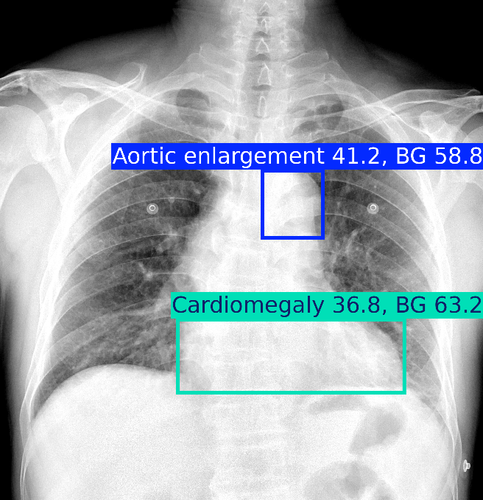}
    \end{subfigure} \hfill
    \begin{subfigure}{0.22\textwidth}
        \centering \includegraphics[width=\linewidth,trim={0 2cm 0 2cm}, clip]{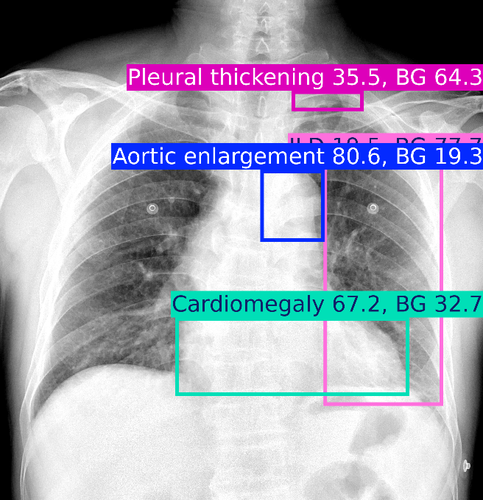}
    \end{subfigure} \hfill
    \begin{subfigure}{0.22\textwidth}
        \centering \includegraphics[width=\linewidth,trim={0 2cm 0 2cm}, clip]{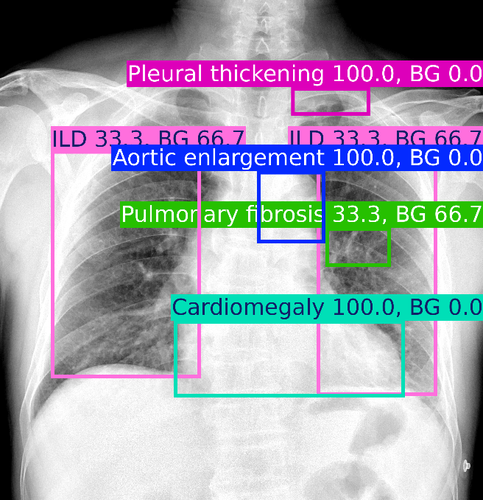}
    \end{subfigure}

    \smallskip
    \begin{subfigure}{0.22\textwidth}
        \centering \includegraphics[width=\linewidth]{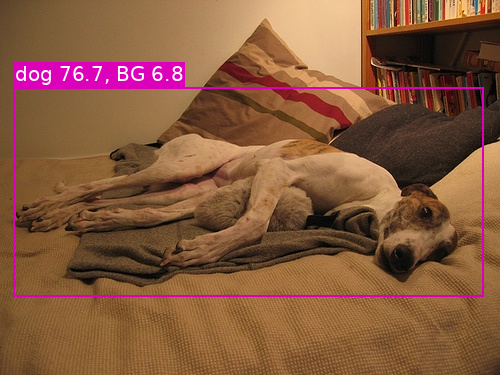}
    \end{subfigure} \hfill
    \begin{subfigure}{0.22\textwidth}
        \centering \includegraphics[width=\linewidth]{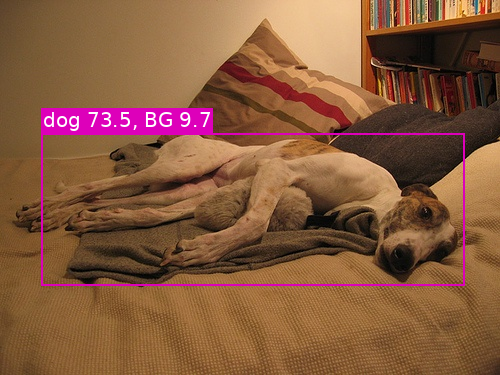}
    \end{subfigure} \hfill
    \begin{subfigure}{0.22\textwidth}
        \centering \includegraphics[width=\linewidth]{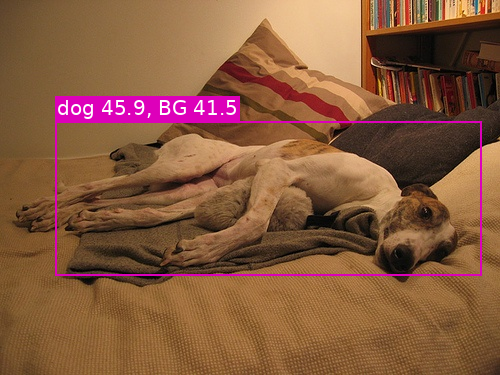}
    \end{subfigure} \hfill
    \begin{subfigure}{0.22\textwidth}
        \centering \includegraphics[width=\linewidth]{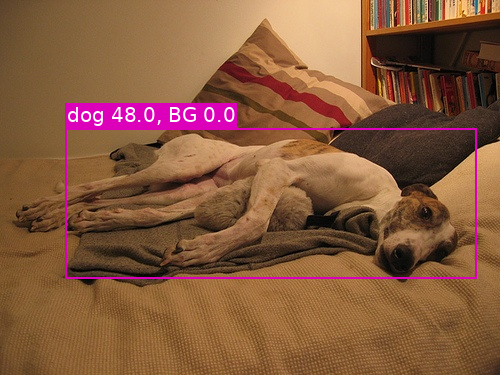}
    \end{subfigure}
    
    \smallskip
    \begin{subfigure}{0.22\textwidth}
        \centering \includegraphics[width=\linewidth,trim={0 7cm 0 2cm},clip]{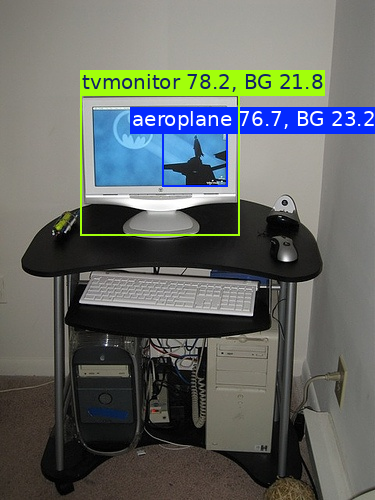}
        \caption{YOLOX}
    \end{subfigure} \hfill
    \begin{subfigure}{0.22\textwidth}
        \centering \includegraphics[width=\linewidth,trim={0 7cm 0 2cm},clip]{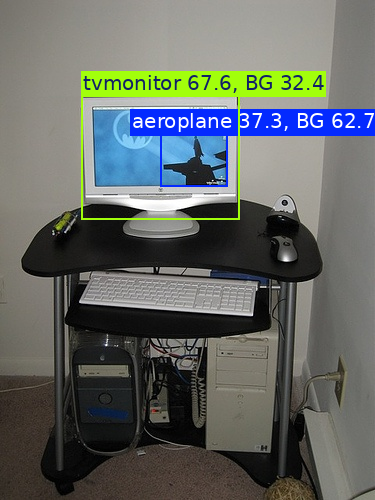}
        \caption{YOLOX-G}
    \end{subfigure} \hfill
    \begin{subfigure}{0.22\textwidth}
        \centering \includegraphics[width=\linewidth,trim={0 7cm 0 2cm},clip]{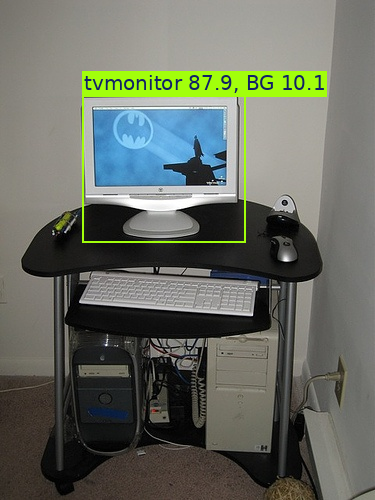}
        \caption{YOLOX-G$\dag$ (ours)}
    \end{subfigure} \hfill
    \begin{subfigure}{0.22\textwidth}
        \centering \includegraphics[width=\linewidth,trim={0 7cm 0 2cm},clip]{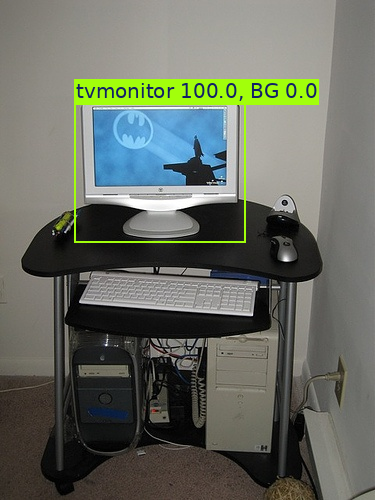}
        \caption{Annotators' agreement}
    \end{subfigure}

    \caption{Comparison of the detection results of (a) YOLOX, (b) YOLOX-G, and (c) YOLOX-G$\dag$ where the labels above each box are the predicted class, its probability, and the background (BG) class's probability. (d): The annotators' agreement obtained from the pre-processing module where the labels above each box represent the percentage of annotators agreeing with the majority class and those not annotating any class for the bounding box (\ie background class). Rows 1-2: DX-20; Row 3-4: VBD-CXR; Row 5-6: VOC-MIX.}
    \label{fig:qualitative}
\end{figure*}

\paragraph{\textbf{Quantitative results}}
\Cref{table:quantitative} reports the evaluation metrics introduced in \Cref{subsec:eval} for the deterministic object detectors, their probabilistic counterpart trained via NLL (*-G), and our proposed method (*-G$\dag$). 
We cannot report the LUE for the deterministic object detectors since they predict bounding boxes as point estimates. However, to perform the matching step during evaluation (in \Cref{subsec:eval}), we assume a constant bounding box variance for the deterministic object detectors.

\Cref{table:quantitative} shows that the deep ensemble models obtain the best mean calibration error in most of the experiment settings. However, the training and inference costs of these models increase linearly with the number of annotators, making them not scalable in real-world applications. In contrast, the models calibrated with our proposed framework perform on par against the deep ensemble models, with the largest mean calibration error difference of 3.1 only.
Additionally, our results highlight the flexibility of the proposed calibration framework in adapting to different object detection architectures for medical images (\ie DX-20 and VBD-CXR) while consistently outperforming other alternatives. Incorporating our calibration method generally leads to improvement across all metrics 
where our gains on the mean metric range from $1.8$ to $6.8$ when compared to training Gaussian object detectors with the NLL loss. 

Furthermore, our method is robust to natural images and a higher number of annotators per image (\ie 25), as demonstrated using the VOC-MIX dataset. We observe performance improvement with the proposed YOLOX-G$\dag$ achieving a mean calibration error of $32.7$. Besides, the proposed method reduces the mean calibration error by $2.6-4.5$ from Gaussian object detectors trained with the NLL loss on the VOC-MIX dataset. This exemplifies our method's superior performance and generalizability to more annotators and object detection in natural images.

Lastly, we note that other calibration metrics (\eg, D-ECE, LaECE, \etc) are not directly applicable to our problem setup, as they have different calibration objectives, are designed for deterministic object detectors, require ground truth annotations, and do not calibrate to predicted bounding box distributions. However, by using majority voting to obtain reference ground truth, we can report the results of additional calibration metrics and methods to supplement our analysis. 
Since the DX-20 and VBD-CXR datasets only contain annotations from multiple annotators (no well-defined ground truth), we use majority voting on the annotation clusters obtained from \Cref{subsec:preprocess} to generate a consensus annotation as a proxy ground truth for evaluation. We note that many annotation clusters in the DX-20 dataset are dropped when the majority class is the `background' class or unable to achieve majority agreement due to low inter-annotator agreement. For the VOC-MIX dataset, we use the original ground truth of the Pascal VOC \cite{pascal-voc-2007} dataset to compute the calibration metrics.

\Cref{table:stand_calib} reports the calibration evaluation metrics LaECE$_0$ and LaACE$_0$ of the isotonic regression (*-IR) and Platt scaling (*-PS) calibration methods used by \citet{kuzucu2024calibration}, with the calibration objective explicitly optimized to reduce LaECE$_0$ and LaACE$_0$. Unsurprisingly, the IR and PS calibration methods achieve the best calibration results. Despite the difference in the calibration objective, our proposed method is still able to improve both the LaECE$_0$ and LaACE$_0$ metrics when compared to the uncalibrated deterministic object detector baselines, while having similar performance to the deep ensemble models.

\begin{figure}[tb]
    \centering
    \begin{subfigure}{0.47\linewidth}
        \centering \includegraphics[width=\linewidth]{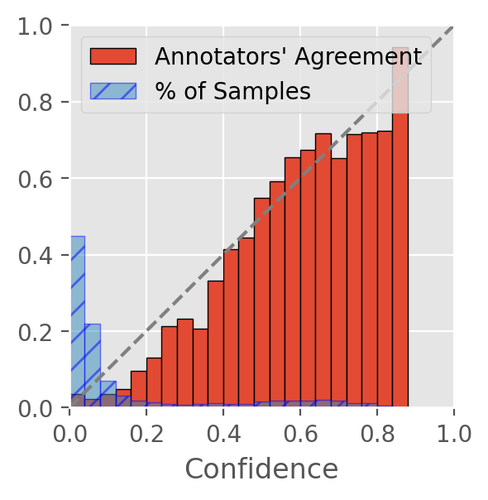}
        \caption{YOLOX-G}
    \end{subfigure}
    \hfill
    \begin{subfigure}{0.47\linewidth}
        \centering \includegraphics[width=\linewidth]{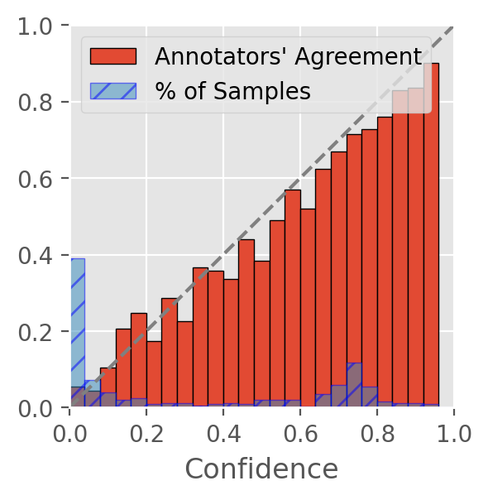}
        \caption{YOLOX-G$\dag$ (ours)}
    \end{subfigure}
    \caption{The reliability diagrams of YOLOX on VOC-MIX. The annotators' agreement refers to the percentage of annotators annotating the same class label as the predicted class label while the \% of samples is the proportion of predictions in each bin.}
    \label{fig:rel_diagram}
\end{figure}

\paragraph{\textbf{Qualitative results}}
\Cref{fig:qualitative} visualizes the detection results of YOLOX, YOLOX-G, and YOLOX-G$\dag$ on the test datasets. Since the display of full predicted distributions will overcrowd the images, we only visualize the highest class probability and the probability of the background class for each predicted object. In \Cref{fig:qualitative}, we observe that the detection results from YOLOX-G$\dag$ are more similar to the annotators' agreement (obtained from the pre-processing module) than YOLOX and YOLOX-G. 
Moreover, \Cref{fig:rel_diagram} presents the reliability diagrams of YOLOX-G and YOLOX-G$\dag$ on VOC-MIX and shows that our method improves the calibration quality towards the reference line (\ie dashed line).
We refer readers to \ref{appx:add_qualitative} for the reliability diagrams of all other experiments.

\begin{table*}[tb]
    \centering\setlength{\tabcolsep}{4pt}
    \caption{Detection accuracy metrics. Ground truth of DX-20 and VBD-CXR are approximated by majority voting, while we use the original ground truth of the Pascal VOC dataset for VOC-MIX. *AP$^{.4}$ is the evaluation metric of the private test set of VBD-CXR, obtained via server-side submission on Kaggle. Higher is better for all metrics except LRP, and bold indicates the best results.}
    \resizebox{\linewidth}{!}{
    \begin{tabular}{l|ccccc|cccccc|ccccc}
        \hline
        \multirow{2}{*}{\textbf{Model}} & \multicolumn{5}{c|}{\textbf{DX-20}} & \multicolumn{6}{c|}{\textbf{VBD-CXR}} & \multicolumn{5}{c}{\textbf{VOC-MIX}} \\
          & AP$^{.5}$ & AP$^{.75}$ & AP$^{.5:.95}$ & LRP$\downarrow$ & PDQ & *AP$^{.4}$& AP$^{.5}$ & AP$^{.75}$ & AP$^{.5:.95}$ & LRP$\downarrow$ & PDQ & AP$^{.5}$ & AP$^{.75}$ & AP$^{.5:.95}$ & LRP$\downarrow$ & PDQ \\
         \hline \hline
         FRCNN & \textbf{25.9} & \textbf{6.7} & \textbf{10.3} & 91.0 & - & \textbf{17.0} & \textbf{34.9} & \textbf{17.0} & 17.6 & \textbf{84.9} & - & 82.0 & 59.5 & 53.7 & \textbf{54.6} & - \\
         FRCNN-DE & \textbf{25.9} & 6.1 & 9.8 & \textbf{90.7} & \textbf{9.33} & 16.8 & 34.3 & 16.1 & \textbf{17.7} & \textbf{84.9} & \textbf{5.25} & \textbf{82.7} & \textbf{60.2} & \textbf{54.0} & \textbf{54.6} & \textbf{12.02} \\
         FRCNN-G & 23.9 & 5.8 & 9.7 & 91.4 & 5.76 & 16.5 & 34.5 & 15.8 & \textbf{17.7} & 85.1 & 5.05 & 82.1 & 59.1 & 53.0 & 54.9 & 6.48 \\
         FRCNN-G$\dag$ (ours) & 25.8 & 6.1 & 10.1 & \textbf{90.7} & 8.30 & 16.2& 33.6 & 16.0 & 17.1 & 85.4 & 4.94 & 81.7 & 59.3 & 53.2 & 55.8 & 11.53 \\
         \hline
         RetinaNet & \textbf{23.7} & \textbf{6.9} & \textbf{10.2} & 91.2 & - & 14.9 & 32.2 & 14.9 & 17.0 & 86.4 & - & \textbf{80.9} & 59.6 & 54.0 & 54.5 & - \\
         RetinaNet-DE & 23.5 & 6.3 & 9.7 & \textbf{90.7} & \textbf{2.16} & \textbf{15.2} & \textbf{33.7} & \textbf{15.3} & \textbf{17.3} & \textbf{85.4} & \textbf{2.97} & 80.8 & \textbf{60.1} & \textbf{54.3} &\textbf{54.0} & \textbf{9.96} \\
         RetinaNet-G & 20.8 & 5.8 & 8.8 & 92.1 & 1.15 & 14.3 & 31.3 & 15.2 & 16.7 & 86.0 & 1.02 & 78.8 & 58.4 & 53.2 & 54.3 & 7.01 \\
         RetinaNet-G$\dag$ (ours) & 23.4 & 5.7 & 9.4 & 91.6 & 1.02 & 14.8 & 31.7 & 14.5 & 16.1 & 85.7 & 1.94 & 79.2 & 59.1 & 53.8 & 55.1 & 9.76 \\
         \hline
         YOLOX & \textbf{38.8} & \textbf{16.8} & \textbf{19.1} & \textbf{83.7} & - & 18.1 & 38.9 & 20.5 & 21.0 & 82.7 & - & 81.3 & 66.0 & \textbf{59.0} & 48.9 & - \\
         YOLOX-DE & 38.6 & 16.2 & 18.5 & 83.8 & \textbf{10.25} & \textbf{19.5} & \textbf{42.5} & \textbf{23.7} & 22.5 & \textbf{80.0} & \textbf{11.36} & 81.4 & \textbf{66.3} & 58.6 & \textbf{48.6} & \textbf{15.55} \\
         YOLOX-G & 35.6 & 14.2 & 17.3 & 85.0 & 7.64 & 18.9 & 40.3 & 22.9 & 21.8 & 81.9 & 7.24 & 80.0 & 63.8 & 57.1 & 50.0 & 10.41 \\
         YOLOX-G$\dag$ (ours) & 38.6 & 15.9 & 18.4 & 83.9 & 9.33 & 18.8 & 41.9 & 23.4 & \textbf{22.6} & 81.0 & 9.51 & \textbf{82.1} & 65.9 & 58.0 & 50.7 & 13.78 \\
         \hline
    \end{tabular}
    }
    \label{table:ap_pdq}
\end{table*}

\begin{table*}[tb]
    \centering
    \caption{Calibration evaluation metrics for the ablation study and the total training and inference computational time. The standard deviation of the calibration metrics across the test samples in the DX-20 dataset is shown in brackets.
    TTCL: Train-time calibration loss.
    }
    \small
    \setlength{\tabcolsep}{4pt}
    \resizebox{0.8\linewidth}{!}{
    \begin{tabular}{l|cccc|c|cc}
        \hline
         \multirow{2}{*}{Model} & \multicolumn{5}{c|}{Calibration Metric} & \multicolumn{2}{c}{Time (s)} \\
          & TVD & TVD\textsubscript{FP} & LUE & FNE & Mean & Train & Infer. \\
         \hline \hline
         RetinaNet-G & 44.4 $(\pm12.5)$ & 45.4 $(\pm8.7)$ & 51.7 $(\pm10.7)$ & \textbf{25.8} $\mathbf{(\pm12.9)}$ & 42.3 $(\pm6.8)$ & 11005.0 & 26.2\\
         \hline
          + TTCL & 42.1 $(\pm12.1)$ & 30.8 $(\pm11.8)$ & 52.5 $(\pm11.1)$ & 27.0 $(\pm14.0)$ & 38.9 $(\pm7.2)$ & 12229.1 & 26.2\\
          \hline
          + Post-hoc IR & \textbf{38.8} $\mathbf{(\pm14.7)}$ & 42.6 $(\pm13.9)$ & 43.5 $(\pm18.1)$ & 27.5 $(\pm11.6)$ & 38.4 $(\pm7.4)$ & 11006.2 & 29.9 \\
          \hline
          + TTCL \& Post-hoc IR & 42.1 $(\pm12.2)$ & \textbf{28.1} $\mathbf{(\pm10.8)}$ & 50.7 $(\pm10.7)$ & \textbf{25.8} $\mathbf{(\pm13.4)}$ & 37.5 $(\pm6.9)$ & 12230.3 & 29.9 \\
          \hline
         \, + sample weights & 40.4 $(\pm11.6)$ & 29.3 $(\pm10.3)$ & \textbf{44.7} $\mathbf{(\pm12.7)}$ & 25.9 $(\pm13.3)$ & \textbf{35.5 $(\pm6.6)$} & 12230.3 & 29.9 \\
         \hline
    \end{tabular}
    }
    \label{table:ablation}
\end{table*}

\paragraph{\textbf{Model interpretation and explainability}}
Our calibrated model can be interpreted by end-users more easily than uncalibrated alternatives, while the calibration metrics can explain the model's overall alignment to the annotators' agreement.
We present an example of interpreting the model output and explaining the calibrated model, specifically, the YOLOX-G$\dag$ model on the DX-20 dataset. A prediction of class ``Peri. Radiol.'' with 48.7\% confidence and 32.4\% background probability (\Cref{fig:qualitative} Row 1) can be interpreted as 48.7\% of annotators agreeing with class ``Peri. Radiol.'', 32.4\% believing there is no object in the bounding box, while the remaining 18.9\% annotating other class labels.
Annotators thinking that an object is present would place their bounding boxes within the 67.6\% (the foreground probability) confidence interval of the predicted bounding box Gaussian distribution. 

In addition, as shown in \Cref{table:quantitative}, YOLOX-G$\dag$ leads to a 38.1 TVD and 9.1 TVD\textsubscript{FP} on DX-20. They can be understood as the largest difference between the predicted class probability and the percentage of annotators agreeing on the prediction on average, which are 38.1 and 9.1 for all annotated objects and FP, respectively. Further, an LUE of 47.5 indicates the mean absolute error of the predicted certainty to represent the percentage of annotated bounding boxes lying within the predicted bounding box interval, while an FNE of 23.5 can be interpreted as 23.5\% of annotators would annotate an object when the model missed it.

\paragraph{\textbf{Detection accuracy is uncompromised by calibration}}
Typically, model performance in terms of accuracy and calibration are both important, \ie, the accuracy of a calibrated model should not deteriorate significantly from its uncalibrated version \cite{kuzucu2024calibration}. To analyze the model's detection accuracy, we supplement our calibration metric by computing the mean AP metric under various IoU thresholds (\ie AP$^{.5}$, AP$^{.75}$, AP$^{.5:.95}$), the LRP metric \cite{lrp}, and the PDQ measure \cite{poddef2020hall} which evaluates the detection quality based on spatial and class probabilities. The reference ground truth is obtained similarly to \Cref{table:stand_calib}. Additionally, for the VBD-CXR dataset, we report the AP$^{0.4}$ metric obtained via server-side submission on Kaggle \footnote{\url{https://www.kaggle.com/competitions/vinbigdata-chest-xray-abnormalities-detection}}.

\Cref{table:ap_pdq} presents the AP, LRP, and PDQ results for the test datasets. We observe that training object detectors with our proposed calibration method (*-G$\dag$) does not significantly reduce detection performance. Particularly, the drop in AP$^{.5:.95}$ ranges between $0.2 - 1.2$, LRP only increases by 2.1 at most, while the worst decrease in PDQ is 1.77, when compared to their deterministic or deep ensemble counterparts. 
This validates that our proposed models can achieve similar performance as their counterparts while yielding well-calibrated class confidence scores and bounding box variances that align with the annotators' annotations distribution.

\paragraph{\textbf{Ablation study}}
We study the effects of each component in the proposed framework by analyzing RetinaNet-G$\dag$, as it leads to the largest reduction in the mean calibration error on the DX-20 dataset. 
As shown in \Cref{table:ablation}, we first add the train-time calibration loss to the RetinaNet-G model, which improves the mean evaluation metric by 3.4. Supervising the model directly with the annotators' agreement significantly improves our model confidence estimates for FP by 14.6 TVD\textsubscript{FP} while only incurring 1224.1 seconds total training time or 0.03 seconds of additional training iteration time. Besides, we incorporate only the post-hoc IR calibrator to the RetinaNet-G model, which leads to a notable drop in LUE of 8.2 compared to the RetinaNet-G model. The post-hoc calibrator improves the mean calibration error of the RetinaNet-G model by 3.9 and takes only 1.2 and 3.7 additional seconds for the IR model training and inference time, respectively.

Then, implementing both train-time and post-hoc calibration further improves the mean metric to 37.5, while using sample weights when training the IR model brings about an extra 2.0 improvement, 
demonstrating the capability of our method in capturing and aligning the model's uncertainty output to the annotators' agreement. 
In short, by utilizing both the train-time calibration loss and the IR calibrator trained with sample weights, we can enhance the mean calibration metric by 6.8 with minimal computational overhead.

\section{Conclusion}
\label{sec:conclusion}
We presented a novel and interpretable approach for calibrating probabilistic object detectors by aligning the model's uncertainty outputs to the annotators' annotation distributions. 
It enables model training and calibration evaluation without requiring ground truth annotations, which can be unavailable for ambiguous objects. 
To achieve this, we introduced four evaluation metrics 
and proposed a calibration method that incorporates a train-time calibration loss and a post-hoc isotonic regression calibrator. The proposed method is generalizable to medical and natural image domains and different types of object detector architectures.
Our results showed that the proposed calibration method achieves the best performance on the calibration metrics without compromising detection accuracy and is computationally efficient.

Despite the promising results, our study has limitations in that our calibration metrics, specifically the localization metrics, cannot be directly compared across architectures. Further, we also assume that the annotators' uncertainty in terms of class labels and bounding boxes is linearly coupled. This assumption may be true in certain cases, where the class label is related to the object location (\eg, in the dentistry domain, the severity grades of dental caries are related to its extent or location in the tooth), but they are often orthogonal in practice. Future work should improve the formulation of the calibration error by considering the localization uncertainty without involving the classification uncertainty.

\section*{CRedit authorship contribution statement}
\textbf{Zhi Qin Tan:} Data curation, formal analysis, methodology, software, validation, visualization, writing - original draft.
\textbf{Owen Addison:} Conceptualization, funding acquisition, supervision, writing - review \& editing.
\textbf{Yunpeng Li:} Conceptualization, funding acquisition, project administration, supervision, writing - review \& editing.

\section*{Declaration of competing interest}
The authors declare that they have no known competing financial interests or personal relationships that could have appeared to influence the work reported in this paper.

\section*{Acknowledgments}
Z.Q. T. acknowledges the support of an ICASE studentship from the EPSRC. The authors acknowledge the support of the National Institute for Health and Care Research (NIHR) through grant NIHR204566.

\clearpage \onecolumn \begin{center} {\Large \textbf{Supplementary Material} \par} \vspace{1.5em} \end{center}
\appendix
This supplementary material provides further details about the dataset used in this work in \ref{appx:dataset},
while \ref{appx:add_qualitative} presents additional reliability diagrams, showcasing our calibrated probabilistic object detector's predictive uncertainty in alignment with the annotators' agreement.

\section{Dataset Details}
\label{appx:dataset}
\subsection{DX-20 Dataset}
\begin{figure}[h]
    \centering
    \includegraphics[width=0.75\linewidth]{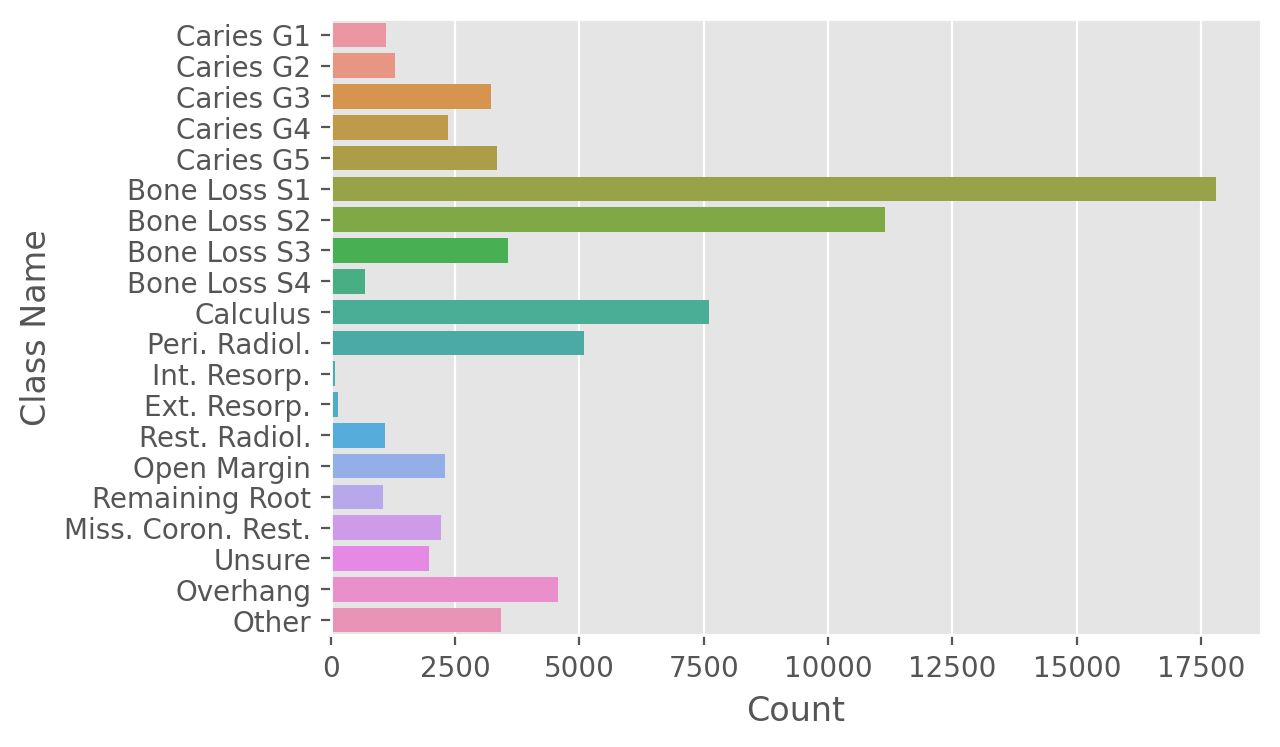}
    \caption{Bar chart of the class label frequency for all annotations of the DX-20 dataset. Caries G1 - G4 and Bone Loss S1 - S4 represent different grades or severity of dental caries and bone loss, respectively.}
    \label{fig:dx20_class}
\end{figure}

Our in-house dental radiograph dataset, DX-20, contains 3,755 bitewing and periapical radiographs collected from 3 different dental practices. 28 specialist dentists with various levels of clinical experience annotated the images with 20 classes of dental diseases (\eg five grades of dental caries, four stages of periodontal bone loss, calculus, \etc). Each image was randomly assigned to 5 different annotators for annotations. 
Annotators labeled bounding boxes around the disease regions and could assign the label ``Unsure'' in case of ambiguity (\eg low image quality). The distribution of the class labels is shown in \Cref{fig:dx20_class}, and we observe significant class imbalance. 

Besides that, an experienced dentist manually selects 600 images with equal proportions from the different dental practices and similar percentages of image difficulties as the testing dataset. In summary, the DX-20 dataset consists of 3,755 images with 467 healthy radiographs (\ie no annotations from any annotator), while the testing set contains 600 images with 101 being healthy radiographs.

\subsection{VBD-CXR Dataset}
VBD-CXR \cite{vindr2022} is a thoracic abnormalities detection dataset with chest (postero-anterior) radiographs, consisting of 15,000 images. The images are annotated by 17 experienced radiologists with 14 classes of abnormalities such as aortic enlargement, atelectasis, calcification, cardiomegaly, \etc. The radiologists are randomly assigned such that there are at least three annotators per image. If no abnormality is found, the radiologist will annotate it with a `no finding` class. Lastly, since the test dataset is not publicly available, we perform a 9:1 split, ensuring similar distribution of object classes on the 15,000 images to obtain our training and testing datasets.

\subsection{VOC-MIX Dataset}
VOC-MIX \cite{tan2024bdc} is a synthetic crowdsourced dataset based on the Pascal VOC \cite{pascal-voc-2007} dataset (object detection in natural images), consisting of 5,011 training and 4,952 testing images. In the VOC-MIX dataset, 25 annotators were simulated to annotate 20 object classes of the Pascal VOC dataset, where 5 annotators have a high annotating accuracy (74\%) and the remaining 20 have an average annotating accuracy (55\%). All annotators annotated every image in the dataset, resulting in 25 distinct sets of annotations.

\clearpage

\section{Additional Reliability Diagrams}
\label{appx:add_qualitative}
\begin{figure*}[bp]
    \centering
    \begin{minipage}{0.48\textwidth}\centering
        FRCNN-G
    \end{minipage}\hfill
    \begin{minipage}{0.48\textwidth}\centering
        FRCNN-G$\dag$
    \end{minipage}
    
    \begin{minipage}{0.48\textwidth}\centering
        \begin{subfigure}{0.49\linewidth}
            \centering \includegraphics[width=\linewidth]{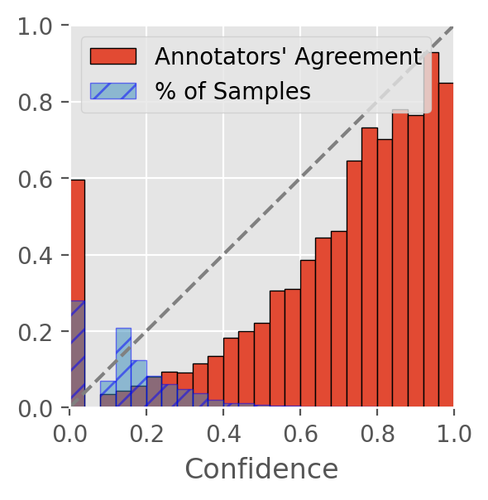}
        \end{subfigure} \hfill
        \begin{subfigure}{0.49\linewidth}
            \centering \includegraphics[width=\linewidth]{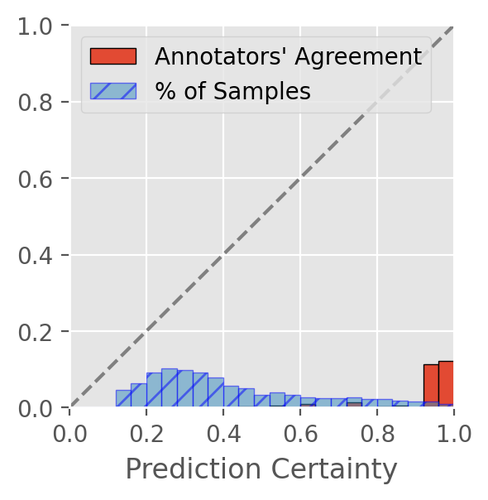}
        \end{subfigure}
    \end{minipage}
    \hfill
    \begin{minipage}{0.48\textwidth}\centering
        \begin{subfigure}{0.49\linewidth}
            \centering \includegraphics[width=\linewidth]{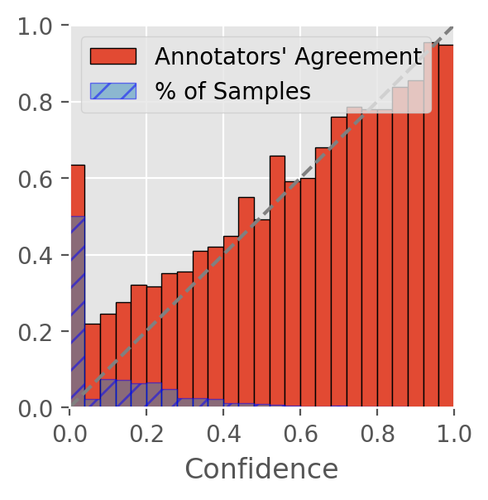}
        \end{subfigure} \hfill
        \begin{subfigure}{0.49\linewidth}
            \centering \includegraphics[width=\linewidth]{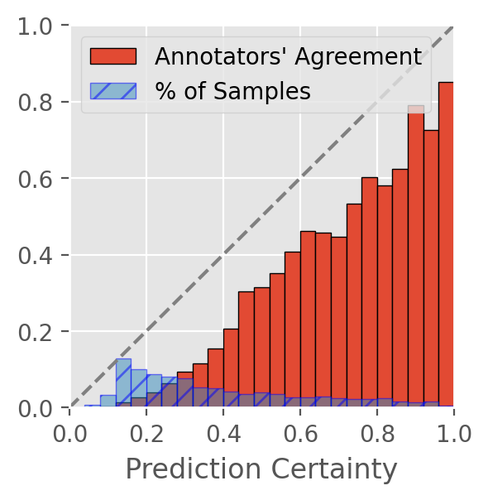}
        \end{subfigure}
    \end{minipage}
    
    \medskip
    \begin{minipage}{0.48\textwidth}\centering
        RetinaNet-G
    \end{minipage}\hfill
    \begin{minipage}{0.48\textwidth}\centering
        RetinaNet-G$\dag$
    \end{minipage}
    
    \begin{minipage}{0.48\textwidth}\centering
        \begin{subfigure}{0.49\linewidth}
            \centering \includegraphics[width=\linewidth]{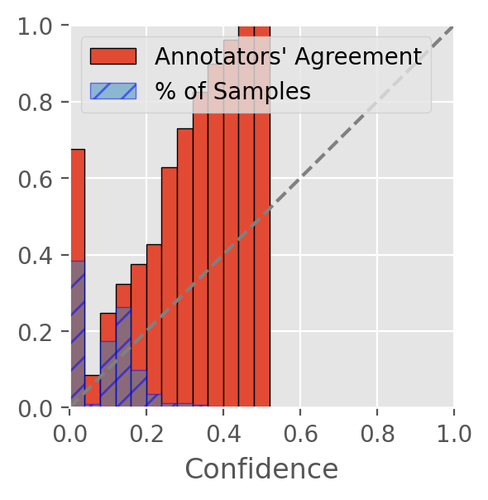}
        \end{subfigure} \hfill
        \begin{subfigure}{0.49\linewidth}
            \centering \includegraphics[width=\linewidth]{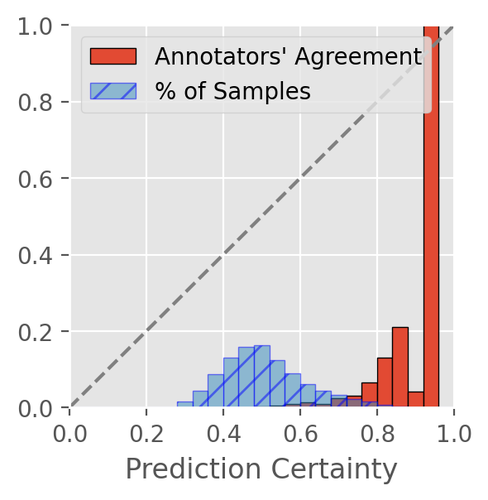}
        \end{subfigure}
    \end{minipage}
    \hfill
    \begin{minipage}{0.48\textwidth}\centering
        \begin{subfigure}{0.49\linewidth}
            \centering \includegraphics[width=\linewidth]{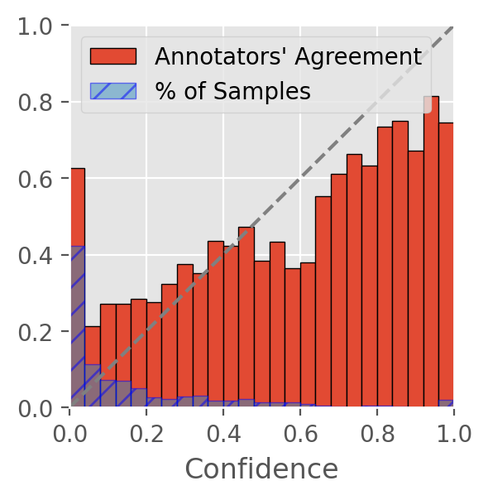}
        \end{subfigure} \hfill
        \begin{subfigure}{0.49\linewidth}
            \centering \includegraphics[width=\linewidth]{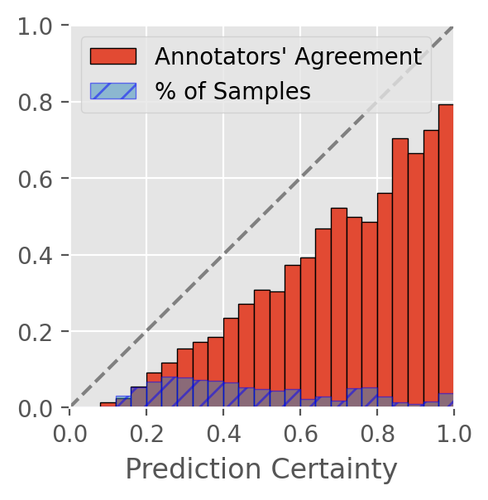}
        \end{subfigure}
    \end{minipage}
        
    \medskip
    \begin{minipage}{0.48\textwidth}\centering
        YOLOX-G
    \end{minipage}\hfill
    \begin{minipage}{0.48\textwidth}\centering
        YOLOX-G$\dag$
    \end{minipage}
    
    \begin{minipage}{0.48\textwidth}\centering\captionsetup[sub]{font=small}
        \begin{subfigure}{0.49\linewidth}
            \centering \includegraphics[width=\linewidth]{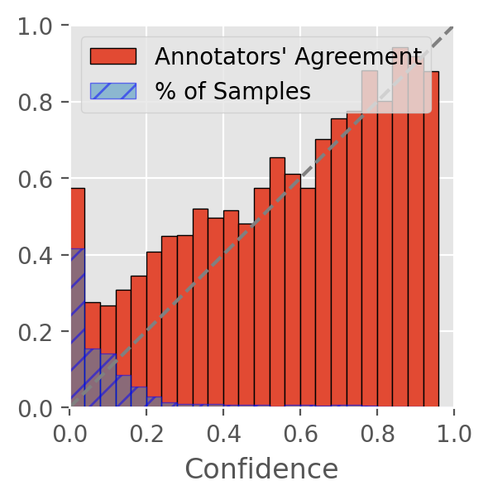}
            \caption{Class Label}
        \end{subfigure} \hfill
        \begin{subfigure}{0.49\linewidth}
            \centering \includegraphics[width=\linewidth]{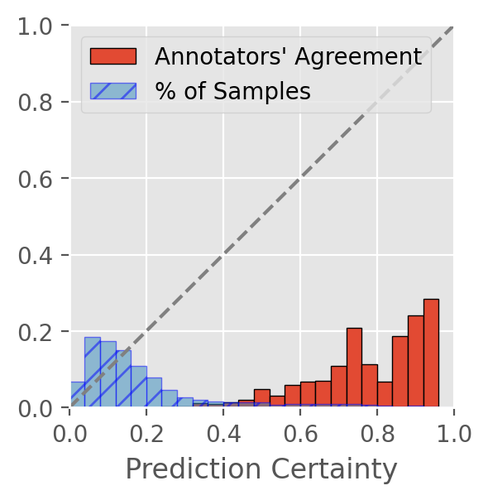}
            \caption{Bounding Box}
        \end{subfigure}
    \end{minipage}
    \hfill
    \begin{minipage}{0.48\textwidth}\centering\captionsetup[sub]{font=small}
        \begin{subfigure}{0.49\linewidth}
            \centering \includegraphics[width=\linewidth]{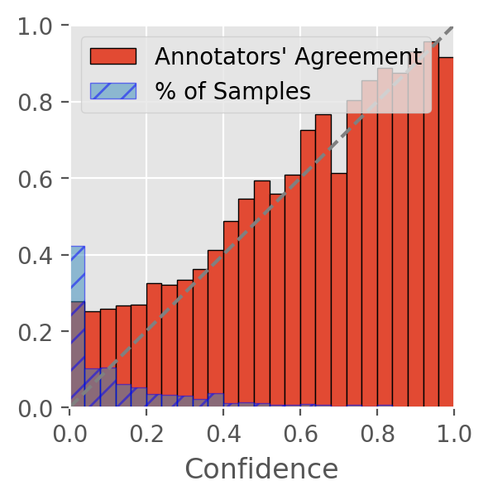}
            \caption{Class Label}
        \end{subfigure} \hfill
        \begin{subfigure}{0.49\linewidth}
            \centering \includegraphics[width=\linewidth]{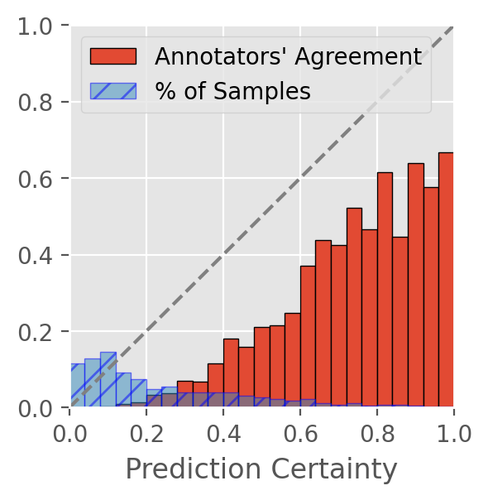}
            \caption{Bounding Box}
        \end{subfigure}
    \end{minipage}
    \caption{The reliability diagram of probabilistic object detectors trained with the NLL loss (*-G) and our proposed calibration method (*-G$\dag$) on the DX-20 dataset. 
    }
    \label{fig:rel_diag_dx20}
\end{figure*}

\begin{figure*}[htbp]
    \centering
    \begin{minipage}{0.48\textwidth}\centering
        FRCNN-G
    \end{minipage}\hfill
    \begin{minipage}{0.48\textwidth}\centering
        FRCNN-G$\dag$
    \end{minipage}
    
    \begin{minipage}{0.48\textwidth}\centering
        \begin{subfigure}{0.49\linewidth}
            \centering \includegraphics[width=\linewidth]{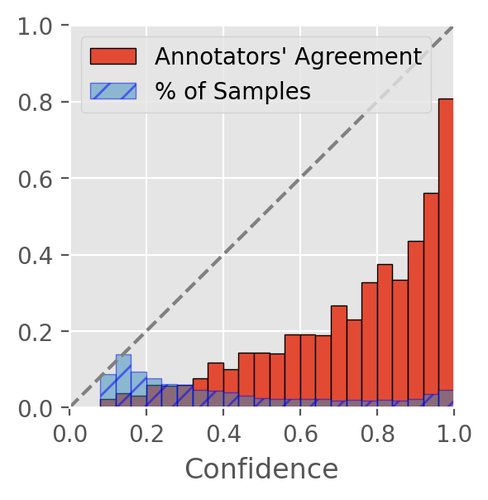}
        \end{subfigure} \hfill
        \begin{subfigure}{0.49\linewidth}
            \centering \includegraphics[width=\linewidth]{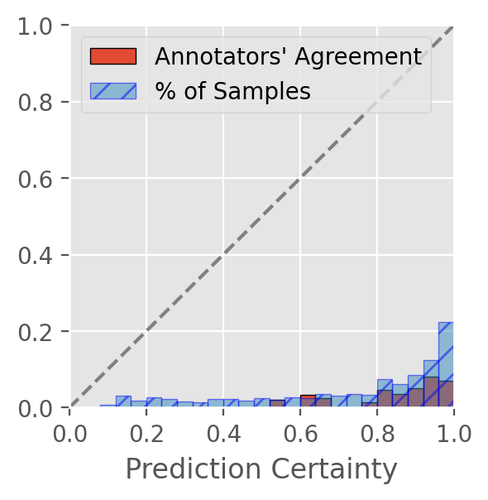}
        \end{subfigure}
    \end{minipage}
    \hfill
    \begin{minipage}{0.48\textwidth}\centering
        \begin{subfigure}{0.49\linewidth}
            \centering \includegraphics[width=\linewidth]{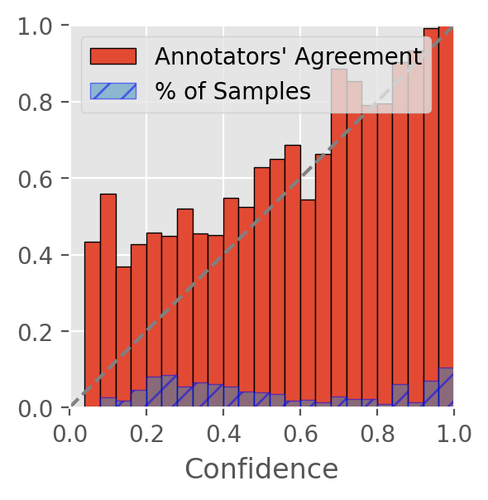}
        \end{subfigure} \hfill
        \begin{subfigure}{0.49\linewidth}
            \centering \includegraphics[width=\linewidth]{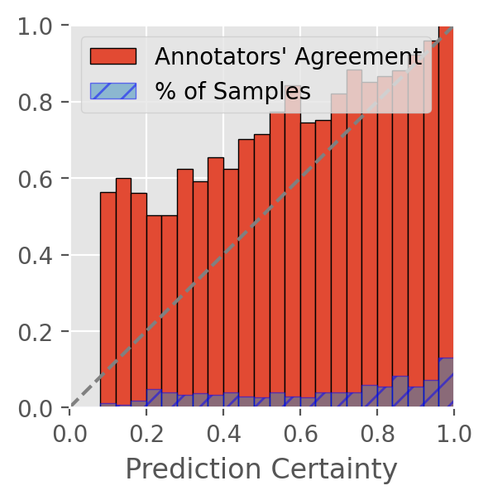}
        \end{subfigure}
    \end{minipage}
    
    \medskip
    \begin{minipage}{0.48\textwidth}\centering
        RetinaNet-G
    \end{minipage}\hfill
    \begin{minipage}{0.48\textwidth}\centering
        RetinaNet-G$\dag$
    \end{minipage}
    
    \begin{minipage}{0.48\textwidth}\centering
        \begin{subfigure}{0.49\linewidth}
            \centering \includegraphics[width=\linewidth]{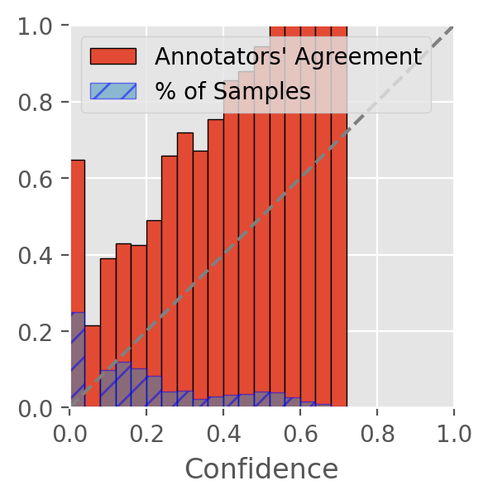}
        \end{subfigure} \hfill
        \begin{subfigure}{0.49\linewidth}
            \centering \includegraphics[width=\linewidth]{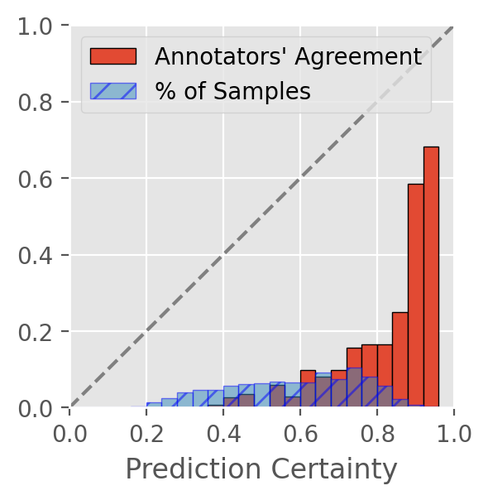}
        \end{subfigure}
    \end{minipage}
    \hfill
    \begin{minipage}{0.48\textwidth}\centering
        \begin{subfigure}{0.49\linewidth}
            \centering \includegraphics[width=\linewidth]{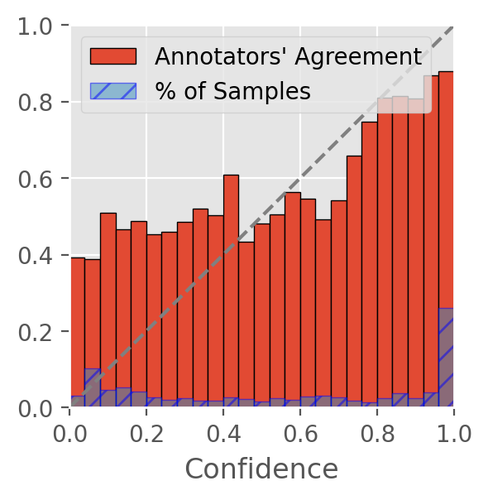}
        \end{subfigure} \hfill
        \begin{subfigure}{0.49\linewidth}
            \centering \includegraphics[width=\linewidth]{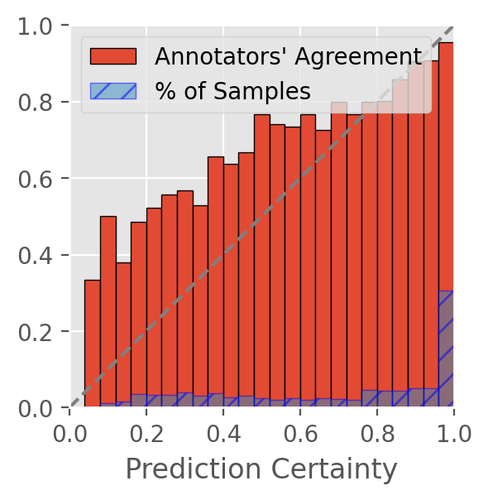}
        \end{subfigure}
    \end{minipage}
        
    \medskip
    \begin{minipage}{0.48\textwidth}\centering
        YOLOX-G
    \end{minipage}\hfill
    \begin{minipage}{0.48\textwidth}\centering
        YOLOX-G$\dag$
    \end{minipage}
    
    \begin{minipage}{0.48\textwidth}\centering\captionsetup[sub]{font=small}
        \begin{subfigure}{0.49\linewidth}
            \centering \includegraphics[width=\linewidth]{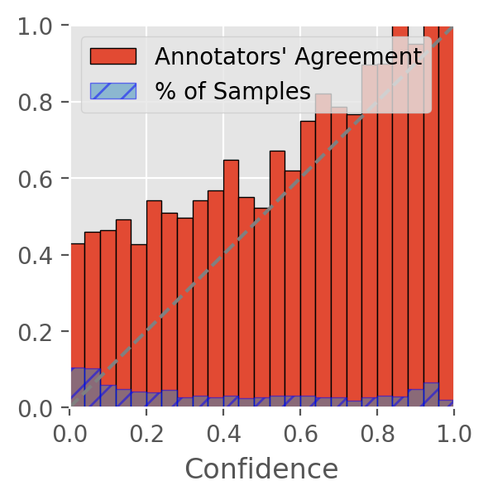}
            \caption{Class Label}
        \end{subfigure} \hfill
        \begin{subfigure}{0.49\linewidth}
            \centering \includegraphics[width=\linewidth]{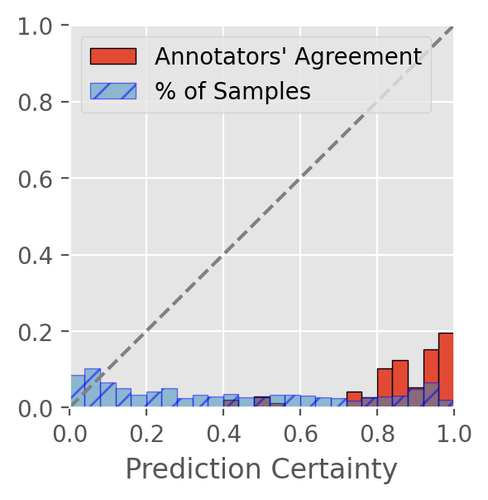}
            \caption{Bounding Box}
        \end{subfigure}
    \end{minipage}
    \hfill
    \begin{minipage}{0.48\textwidth}\centering\captionsetup[sub]{font=small}
        \begin{subfigure}{0.49\linewidth}
            \centering \includegraphics[width=\linewidth]{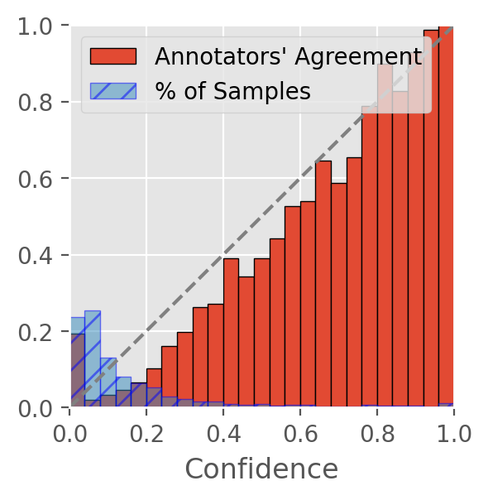}
            \caption{Class Label}
        \end{subfigure} \hfill
        \begin{subfigure}{0.49\linewidth}
            \centering \includegraphics[width=\linewidth]{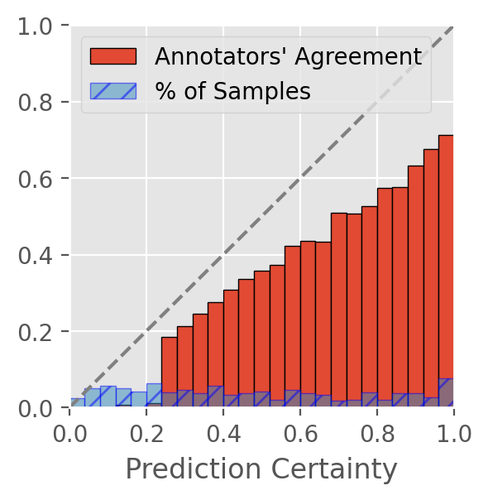}
            \caption{Bounding Box}
        \end{subfigure}
    \end{minipage}
    \caption{The reliability diagram of probabilistic object detectors trained with the NLL loss (*-G) and our proposed calibration method (*-G$\dag$) on the VBD-CXR dataset. 
    }
    \label{fig:rel_diag_vindr}
\end{figure*}

\begin{figure*}[htbp]
    \centering
    \begin{minipage}{0.48\textwidth}\centering
        FRCNN-G
    \end{minipage}\hfill
    \begin{minipage}{0.48\textwidth}\centering
        FRCNN-G$\dag$
    \end{minipage}
    
    \begin{minipage}{0.48\textwidth}\centering
        \begin{subfigure}{0.49\linewidth}
            \centering \includegraphics[width=\linewidth]{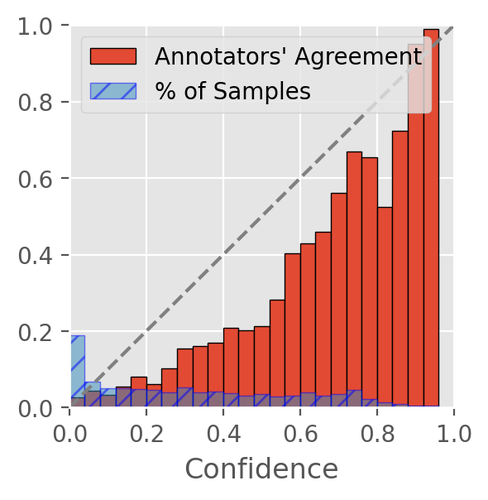}
        \end{subfigure} \hfill
        \begin{subfigure}{0.49\linewidth}
            \centering \includegraphics[width=\linewidth]{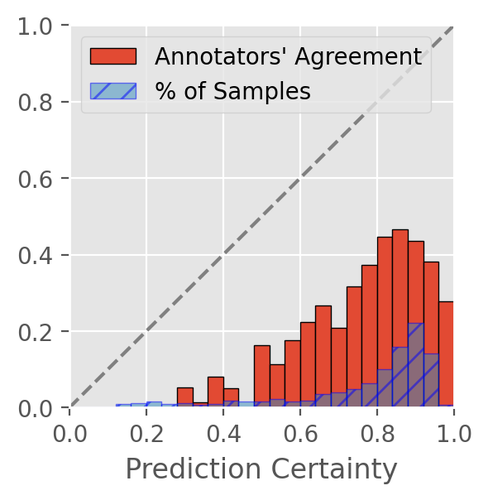}
        \end{subfigure}
    \end{minipage}
    \hfill
    \begin{minipage}{0.48\textwidth}\centering
        \begin{subfigure}{0.49\linewidth}
            \centering \includegraphics[width=\linewidth]{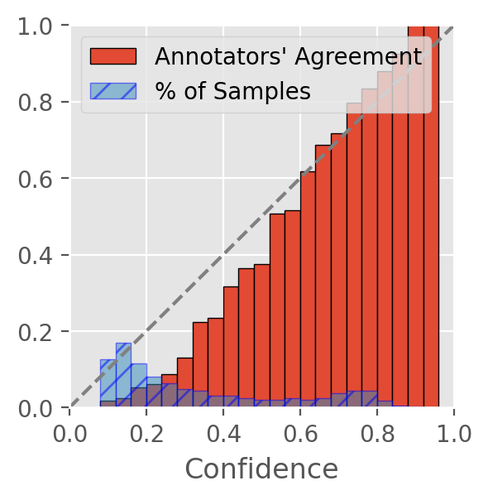}
        \end{subfigure} \hfill
        \begin{subfigure}{0.49\linewidth}
            \centering \includegraphics[width=\linewidth]{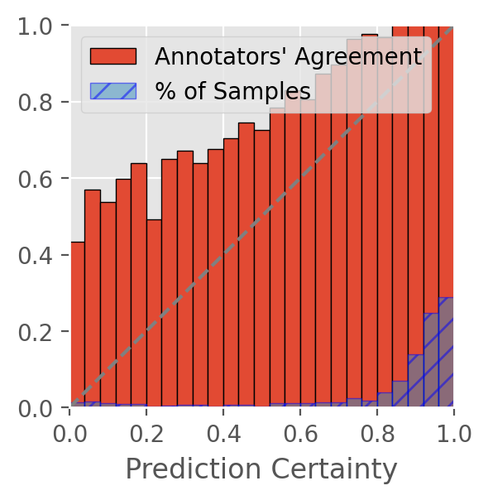}
        \end{subfigure}
    \end{minipage}
    
    \medskip
    \begin{minipage}{0.48\textwidth}\centering
        RetinaNet-G
    \end{minipage}\hfill
    \begin{minipage}{0.48\textwidth}\centering
        RetinaNet-G$\dag$
    \end{minipage}
    
    \begin{minipage}{0.48\textwidth}\centering
        \begin{subfigure}{0.49\linewidth}
            \centering \includegraphics[width=\linewidth]{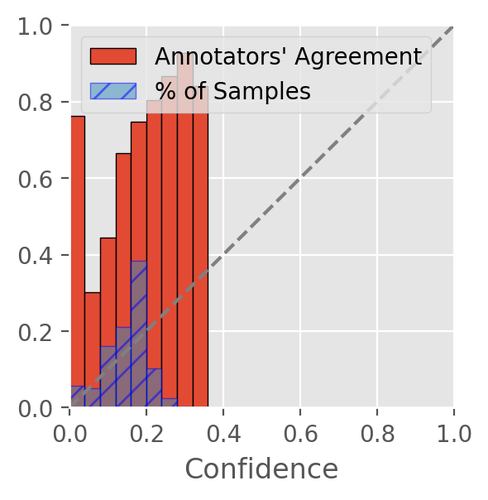}
        \end{subfigure} \hfill
        \begin{subfigure}{0.49\linewidth}
            \centering \includegraphics[width=\linewidth]{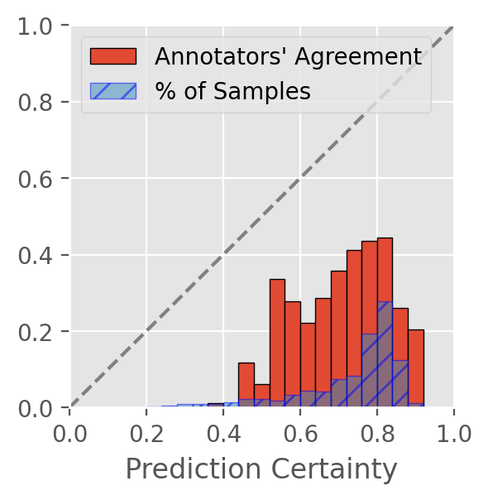}
        \end{subfigure}
    \end{minipage}
    \hfill
    \begin{minipage}{0.48\textwidth}\centering
        \begin{subfigure}{0.49\linewidth}
            \centering \includegraphics[width=\linewidth]{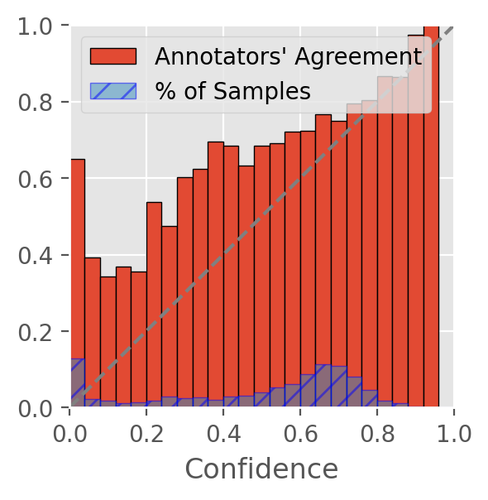}
        \end{subfigure} \hfill
        \begin{subfigure}{0.49\linewidth}
            \centering \includegraphics[width=\linewidth]{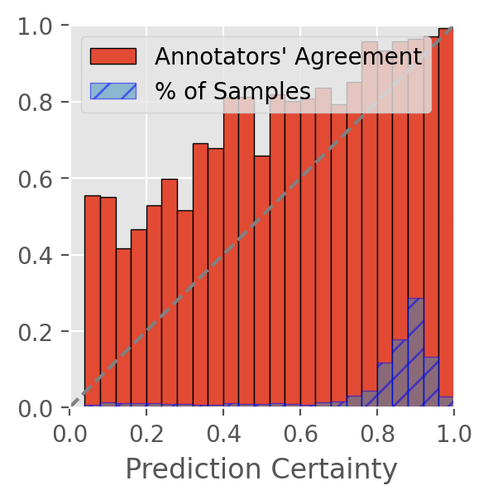}
        \end{subfigure}
    \end{minipage}
        
    \medskip
    \begin{minipage}{0.48\textwidth}\centering
        YOLOX-G
    \end{minipage}\hfill
    \begin{minipage}{0.48\textwidth}\centering
        YOLOX-G$\dag$
    \end{minipage}
    
    \begin{minipage}{0.48\textwidth}\centering\captionsetup[sub]{font=small}
        \begin{subfigure}{0.49\linewidth}
            \centering \includegraphics[width=\linewidth]{images/voc_yolox_nll.png}
            \caption{Class Label}
        \end{subfigure} \hfill
        \begin{subfigure}{0.49\linewidth}
            \centering \includegraphics[width=\linewidth]{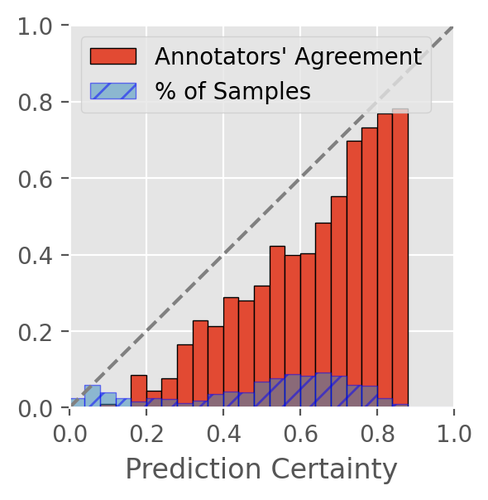}
            \caption{Bounding Box}
        \end{subfigure}
    \end{minipage}
    \hfill
    \begin{minipage}{0.48\textwidth}\centering\captionsetup[sub]{font=small}
        \begin{subfigure}{0.49\linewidth}
            \centering \includegraphics[width=\linewidth]{images/voc_yolox_dmm_ir.png}
            \caption{Class Label}
        \end{subfigure} \hfill
        \begin{subfigure}{0.49\linewidth}
            \centering \includegraphics[width=\linewidth]{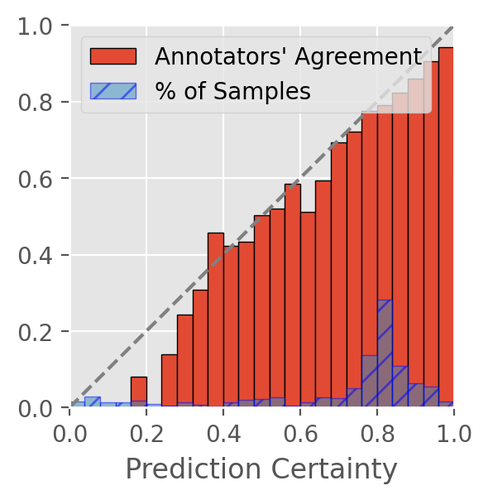}
            \caption{Bounding Box}
        \end{subfigure}
    \end{minipage}
    \caption{The reliability diagram of probabilistic object detectors trained with the NLL loss (*-G) and our proposed calibration method (*-G$\dag$) on the VOC-MIX dataset. 
    }
    \label{fig:rel_diag_voc}
\end{figure*}

\Cref{fig:rel_diag_dx20,fig:rel_diag_vindr,fig:rel_diag_voc} present the reliability diagrams of probabilistic object detectors trained with the NLL loss (*-G) and our proposed calibration method (*-G$\dag$). 
The annotators' agreement in Columns (a) and (c) (\ie class label) refers to the percentage of annotators annotating the same class label as the predicted class label, while in Columns (b) and (d) (\ie bounding box), it indicates the percentage of annotators that annotated their bounding boxes within the predicted bounding box interval calculated with the prediction certainty, $\hat{o}$, as the confidence interval. The bar height of the annotators' agreement is the average percentage in each bin, while the \% of samples is the proportion of predictions in each bin. 

We observe that our proposed calibration method (*-G$\dag$) significantly improves the calibration quality, demonstrated by the closer alignment between the annotators' agreement and the reference line (\ie dashed line) across multiple probabilistic object detectors and datasets.

\clearpage
{
    \small
    \bibliographystyle{elsarticle-num-names}
    \bibliography{references}

\begin{thebibliography}{63}
\expandafter\ifx\csname natexlab\endcsname\relax\def\natexlab#1{#1}\fi
\providecommand{\url}[1]{\texttt{#1}}
\providecommand{\href}[2]{#2}
\providecommand{\path}[1]{#1}
\providecommand{\DOIprefix}{doi:}
\providecommand{\ArXivprefix}{arXiv:}
\providecommand{\URLprefix}{URL: }
\providecommand{\Pubmedprefix}{pmid:}
\providecommand{\doi}[1]{\href{http://dx.doi.org/#1}{\path{#1}}}
\providecommand{\Pubmed}[1]{\href{pmid:#1}{\path{#1}}}
\providecommand{\bibinfo}[2]{#2}
\ifx\xfnm\relax \def\xfnm[#1]{\unskip,\space#1}\fi
\bibitem[{Nguyen et~al.(2020)Nguyen, Nguyen, Nguyen, Elliott, Nguyen, and
  Culliton}]{vindr2022}
\bibinfo{author}{D.~Nguyen}, \bibinfo{author}{D.~B. Nguyen},
  \bibinfo{author}{H.~Q. Nguyen}, \bibinfo{author}{J.~Elliott},
  \bibinfo{author}{N.~T. Nguyen}, \bibinfo{author}{P.~Culliton},
  \bibinfo{title}{{VinBigData} chest x-ray abnormalities detection},
  \bibinfo{howpublished}{\url{https://kaggle.com/competitions/vinbigdata-chest-xray-abnormalities-detection}},
  \bibinfo{year}{2020}. \bibinfo{note}{{Kaggle}}.
\bibitem[{Hamamci et~al.(2021)Hamamci, Er, Simsar, Yuksel, Gultekin, Ozdemir,
  Yang, Li, Pati, Stadlinger et~al.}]{dentex2023}
\bibinfo{author}{I.~E. Hamamci}, \bibinfo{author}{S.~Er},
  \bibinfo{author}{E.~Simsar}, \bibinfo{author}{A.~E. Yuksel},
  \bibinfo{author}{S.~Gultekin}, \bibinfo{author}{S.~D. Ozdemir},
  \bibinfo{author}{K.~Yang}, \bibinfo{author}{H.~B. Li},
  \bibinfo{author}{S.~Pati}, \bibinfo{author}{B.~Stadlinger}, et~al.,
  \bibinfo{title}{{DENTEX}: An abnormal tooth detection with dental enumeration
  and diagnosis benchmark for panoramic x-rays}, \bibinfo{year}{2021}.
  \href{http://arxiv.org/abs/2305.19112}{{\tt arXiv:2305.19112}}.
\bibitem[{Dibo et~al.(2024)Dibo, Galichin, Astashev, Dylov, and
  Rogov}]{deeploc2024}
\bibinfo{author}{R.~Dibo}, \bibinfo{author}{A.~Galichin},
  \bibinfo{author}{P.~Astashev}, \bibinfo{author}{D.~V. Dylov},
  \bibinfo{author}{O.~Y. Rogov},
\newblock \bibinfo{title}{{DeepLOC}: Deep learning-based bone pathology
  localization and classification in wrist x-ray images},
\newblock in: \bibinfo{booktitle}{Proc. Anal. Images Social Netw. Texts},
  \bibinfo{year}{2024}, pp. \bibinfo{pages}{199--211}.
\bibitem[{Everingham et~al.(2010)Everingham, Luc Van~Gool, Winn, and
  Zisserman}]{pascal-voc-2007}
\bibinfo{author}{M.~Everingham}, \bibinfo{author}{C.~K. I.~W. Luc Van~Gool},
  \bibinfo{author}{J.~Winn}, \bibinfo{author}{A.~Zisserman},
\newblock \bibinfo{title}{The {PASCAL} {V}isual {O}bject {C}lasses {C}hallenge
  2007 {(VOC2007)}},
\newblock \bibinfo{journal}{Int. J. Comput. Vis. (IJCV)} \bibinfo{volume}{88}
  (\bibinfo{year}{2010}) \bibinfo{pages}{303--338}.
\bibitem[{Lin et~al.(2014)Lin, Maire, Belongie, Hays, Perona, Ramanan, Dollar,
  and Zitnick}]{cocodataset}
\bibinfo{author}{T.-Y. Lin}, \bibinfo{author}{M.~Maire},
  \bibinfo{author}{S.~Belongie}, \bibinfo{author}{J.~Hays},
  \bibinfo{author}{P.~Perona}, \bibinfo{author}{D.~Ramanan},
  \bibinfo{author}{P.~Dollar}, \bibinfo{author}{L.~Zitnick},
\newblock \bibinfo{title}{Microsoft {COCO}: Common objects in context},
\newblock in: \bibinfo{booktitle}{Proc. Eur. Conf. Comput. Vis. (ECCV)},
  \bibinfo{address}{Zurich, Switzerland}, \bibinfo{year}{2014}, pp.
  \bibinfo{pages}{740--755}.
\bibitem[{Sun et~al.(2020)Sun, Kretzschmar, Dotiwalla, Chouard, Patnaik, Tsui,
  Guo, Zhou, Chai, Caine, Vasudevan, Han, Ngiam, Zhao, Timofeev, Ettinger,
  Krivokon, Gao, Joshi, Zhang, Shlens, Chen, and Anguelov}]{waymo2020}
\bibinfo{author}{P.~Sun}, \bibinfo{author}{H.~Kretzschmar},
  \bibinfo{author}{X.~Dotiwalla}, \bibinfo{author}{A.~Chouard},
  \bibinfo{author}{V.~Patnaik}, \bibinfo{author}{P.~Tsui},
  \bibinfo{author}{J.~Guo}, \bibinfo{author}{Y.~Zhou},
  \bibinfo{author}{Y.~Chai}, \bibinfo{author}{B.~Caine},
  \bibinfo{author}{V.~Vasudevan}, \bibinfo{author}{W.~Han},
  \bibinfo{author}{J.~Ngiam}, \bibinfo{author}{H.~Zhao},
  \bibinfo{author}{A.~Timofeev}, \bibinfo{author}{S.~Ettinger},
  \bibinfo{author}{M.~Krivokon}, \bibinfo{author}{A.~Gao},
  \bibinfo{author}{A.~Joshi}, \bibinfo{author}{Y.~Zhang},
  \bibinfo{author}{J.~Shlens}, \bibinfo{author}{Z.~Chen},
  \bibinfo{author}{D.~Anguelov},
\newblock \bibinfo{title}{Scalability in perception for autonomous driving:
  Waymo open dataset},
\newblock in: \bibinfo{booktitle}{Proc. IEEE Conf. Comput. Vis. Pattern Recog.
  (CVPR)}, \bibinfo{year}{2020}.
\bibitem[{Quinn et~al.(2023)Quinn, Tryposkiadis, Deeks, De~Vet, Mallett,
  Mokkink, Takwoingi, Taylor-Phillips, and Sitch}]{interobsstudy2023}
\bibinfo{author}{L.~Quinn}, \bibinfo{author}{K.~Tryposkiadis},
  \bibinfo{author}{J.~Deeks}, \bibinfo{author}{H.~C. De~Vet},
  \bibinfo{author}{S.~Mallett}, \bibinfo{author}{L.~B. Mokkink},
  \bibinfo{author}{Y.~Takwoingi}, \bibinfo{author}{S.~Taylor-Phillips},
  \bibinfo{author}{A.~Sitch},
\newblock \bibinfo{title}{Interobserver variability studies in diagnostic
  imaging: a methodological systematic review},
\newblock \bibinfo{journal}{Brit. J. Radiol.} \bibinfo{volume}{96}
  (\bibinfo{year}{2023}) \bibinfo{pages}{20220972}.
\bibitem[{Tschirschwitz et~al.(2022)Tschirschwitz, Klemstein, Stein, and
  Rodehorst}]{david_krippendorff2022}
\bibinfo{author}{D.~Tschirschwitz}, \bibinfo{author}{F.~Klemstein},
  \bibinfo{author}{B.~Stein}, \bibinfo{author}{V.~Rodehorst},
\newblock \bibinfo{title}{A dataset for analysing complex document layouts in
  the digital humanities and its evaluation with krippendorff’s alpha},
\newblock in: \bibinfo{booktitle}{Proc. German Conf. Pattern Recog.},
  \bibinfo{address}{Konstanz, Germany}, \bibinfo{year}{2022}, p.
  \bibinfo{pages}{354–374}.
\bibitem[{Hu and Meina(2020)}]{crowdrcnn2020}
\bibinfo{author}{Y.~Hu}, \bibinfo{author}{S.~Meina},
\newblock \bibinfo{title}{Crowd {R-CNN}: An object detection model utilizing
  crowdsourced labels},
\newblock in: \bibinfo{booktitle}{Proc. Int. Conf. Vis. Image Sig. Process.
  (ICVISP)}, \bibinfo{address}{Bangkok, Thailand}, \bibinfo{year}{2020}, pp.
  \bibinfo{pages}{1--7}.
\bibitem[{Le et~al.(2023)Le, Tran, Pham, Nguyen~Trung, Le, and Nguyen}]{Le2023}
\bibinfo{author}{K.~Le}, \bibinfo{author}{T.~Tran}, \bibinfo{author}{H.~Pham},
  \bibinfo{author}{H.~Nguyen~Trung}, \bibinfo{author}{T.~Le},
  \bibinfo{author}{H.~Q. Nguyen},
\newblock \bibinfo{title}{Learning from multiple expert annotators for
  enhancing anomaly detection in medical image analysis},
\newblock \bibinfo{journal}{IEEE Access} \bibinfo{volume}{11}
  (\bibinfo{year}{2023}) \bibinfo{pages}{14105--14114}.
\bibitem[{Tan et~al.(2024)Tan, Isupova, Carneiro, Zhu, and Li}]{tan2024bdc}
\bibinfo{author}{Z.~Q. Tan}, \bibinfo{author}{O.~Isupova},
  \bibinfo{author}{G.~Carneiro}, \bibinfo{author}{X.~Zhu},
  \bibinfo{author}{Y.~Li},
\newblock \bibinfo{title}{Bayesian detector combination for object detection
  with crowdsourced annotations},
\newblock in: \bibinfo{booktitle}{Proc. Eur. Conf. Comput. Vis. (ECCV)},
  \bibinfo{address}{Milan, Italy}, \bibinfo{year}{2024}, pp.
  \bibinfo{pages}{329--346}.
\bibitem[{Kuppers et~al.(2020)Kuppers, Kronenberger, Shantia, and
  Haselhoff}]{d_ece2020}
\bibinfo{author}{F.~Kuppers}, \bibinfo{author}{J.~Kronenberger},
  \bibinfo{author}{A.~Shantia}, \bibinfo{author}{A.~Haselhoff},
\newblock \bibinfo{title}{{Multivariate Confidence Calibration for Object
  Detection}},
\newblock in: \bibinfo{booktitle}{IEEE Conf. Comput. Vis. Pattern Recog. (CVPR)
  Worksh.}, \bibinfo{year}{2020}, pp. \bibinfo{pages}{1322--1330}.
\bibitem[{Oksuz et~al.(2023)Oksuz, Joy, and Dokania}]{la_ece2023}
\bibinfo{author}{K.~Oksuz}, \bibinfo{author}{T.~Joy}, \bibinfo{author}{P.~K.
  Dokania},
\newblock \bibinfo{title}{Towards building self-aware object detectors via
  reliable uncertainty quantification and calibration},
\newblock in: \bibinfo{booktitle}{Proc. IEEE Conf. Comput. Vis. Pattern Recog.
  (CVPR)}, \bibinfo{address}{Vancouver, Canada}, \bibinfo{year}{2023}, pp.
  \bibinfo{pages}{9263--9274}.
\bibitem[{Mehrtash et~al.(2020)Mehrtash, Wells, Tempany, Abolmaesumi, and
  Kapur}]{Mehrtash2020qi}
\bibinfo{author}{A.~Mehrtash}, \bibinfo{author}{W.~M. Wells},
  \bibinfo{author}{C.~M. Tempany}, \bibinfo{author}{P.~Abolmaesumi},
  \bibinfo{author}{T.~Kapur},
\newblock \bibinfo{title}{Confidence calibration and predictive uncertainty
  estimation for deep medical image segmentation},
\newblock \bibinfo{journal}{IEEE Trans. Med. Imag.} \bibinfo{volume}{39}
  (\bibinfo{year}{2020}) \bibinfo{pages}{3868--3878}.
\bibitem[{Chen et~al.(2022)Chen, Wang, Zhang, Fung, Thai, Moore, Mannel, Liu,
  Zheng, and Qiu}]{CHEN2022102444}
\bibinfo{author}{X.~Chen}, \bibinfo{author}{X.~Wang},
  \bibinfo{author}{K.~Zhang}, \bibinfo{author}{K.-M. Fung},
  \bibinfo{author}{T.~C. Thai}, \bibinfo{author}{K.~Moore},
  \bibinfo{author}{R.~S. Mannel}, \bibinfo{author}{H.~Liu},
  \bibinfo{author}{B.~Zheng}, \bibinfo{author}{Y.~Qiu},
\newblock \bibinfo{title}{Recent advances and clinical applications of deep
  learning in medical image analysis},
\newblock \bibinfo{journal}{Med. Image Anal.} \bibinfo{volume}{79}
  (\bibinfo{year}{2022}) \bibinfo{pages}{102444}.
\bibitem[{Gawlikowski et~al.(2023)Gawlikowski, Tassi, Ali, Lee, Humt, Feng,
  Kruspe, Triebel, Jung, Roscher, Shahzad, Yang, Bamler, and
  Zhu}]{Gawlikowski2023}
\bibinfo{author}{J.~Gawlikowski}, \bibinfo{author}{C.~R.~N. Tassi},
  \bibinfo{author}{M.~Ali}, \bibinfo{author}{J.~Lee},
  \bibinfo{author}{M.~Humt}, \bibinfo{author}{J.~Feng},
  \bibinfo{author}{A.~Kruspe}, \bibinfo{author}{R.~Triebel},
  \bibinfo{author}{P.~Jung}, \bibinfo{author}{R.~Roscher},
  \bibinfo{author}{M.~Shahzad}, \bibinfo{author}{W.~Yang},
  \bibinfo{author}{R.~Bamler}, \bibinfo{author}{X.~X. Zhu},
\newblock \bibinfo{title}{A survey of uncertainty in deep neural networks},
\newblock \bibinfo{journal}{Artif. Intell. Rev.} \bibinfo{volume}{56}
  (\bibinfo{year}{2023}) \bibinfo{pages}{1513--1589}.
\bibitem[{Feng et~al.(2019)Feng, Rosenbaum, Glaeser, Timm, and
  Dietmayer}]{dmm2019}
\bibinfo{author}{D.~Feng}, \bibinfo{author}{L.~Rosenbaum},
  \bibinfo{author}{C.~Glaeser}, \bibinfo{author}{F.~Timm},
  \bibinfo{author}{K.~Dietmayer},
\newblock \bibinfo{title}{Can we trust you? on calibration of a probabilistic
  object detector for autonomous driving},
\newblock in: \bibinfo{booktitle}{Proc. IEEE/RSJ Int. Conf. Intell. Robots
  Syst. (IROS)}, \bibinfo{address}{Macao, China}, \bibinfo{year}{2019}.
\bibitem[{Nowak et~al.(2025)Nowak, Cyranka, Maslany, Kostuch, Derbisz,
  Komorkiewicz, Siwck, Wojcik, Marchewka, and Skruch}]{nowak_cvprw2025}
\bibinfo{author}{M.~K. Nowak}, \bibinfo{author}{J.~Cyranka},
  \bibinfo{author}{N.~Maslany}, \bibinfo{author}{A.~Kostuch},
  \bibinfo{author}{J.~Derbisz}, \bibinfo{author}{M.~Komorkiewicz},
  \bibinfo{author}{P.~Siwck}, \bibinfo{author}{M.~J. Wojcik},
  \bibinfo{author}{D.~Marchewka}, \bibinfo{author}{P.~Skruch},
\newblock \bibinfo{title}{{ How Much Noise is There in Labels Generated by
  Humans? A Method to Validate Automatically Generated Bounding Boxes }},
\newblock in: \bibinfo{booktitle}{IEEE Conf. Comput. Vis. Pattern Recog. (CVPR)
  Worksh.}, \bibinfo{address}{Los Alamitos, CA, USA}, \bibinfo{year}{2025}, pp.
  \bibinfo{pages}{4371--4380}.
\bibitem[{Tschirschwitz and Rodehorst(2025)}]{tschir_wacv2025}
\bibinfo{author}{D.~Tschirschwitz}, \bibinfo{author}{V.~Rodehorst},
\newblock \bibinfo{title}{{ Label Convergence: Defining an Upper Performance
  Bound in Object Recognition Through Contradictory Annotations }},
\newblock in: \bibinfo{booktitle}{Proc. IEEE Winter Conf. Appl. Comput. Vis.
  (WACV)}, \bibinfo{address}{Tucson, AZ, USA}, \bibinfo{year}{2025}, pp.
  \bibinfo{pages}{6848--6857}.
\bibitem[{Zhou et~al.(2024)Zhou, Li, Li, Huang, Nie, Liu, Gao, Wang, Heng, and
  Chen}]{Zhou_2024_BMVC}
\bibinfo{author}{D.~Zhou}, \bibinfo{author}{J.~Li}, \bibinfo{author}{J.~Li},
  \bibinfo{author}{J.~Huang}, \bibinfo{author}{Q.~Nie},
  \bibinfo{author}{Y.~Liu}, \bibinfo{author}{B.-B. Gao},
  \bibinfo{author}{Q.~Wang}, \bibinfo{author}{P.-A. Heng},
  \bibinfo{author}{G.~Chen},
\newblock \bibinfo{title}{Distribution-aware calibration for object detection
  with noisy bounding boxes},
\newblock in: \bibinfo{booktitle}{Proc. Brit. Mach. Vis. Conf. (BMVC)},
  \bibinfo{address}{Glasgow, UK}, \bibinfo{year}{2024}.
\bibitem[{Murrugarra-Llerena and Jung(2025)}]{murru_wacv2025}
\bibinfo{author}{J.~Murrugarra-Llerena}, \bibinfo{author}{C.~R. Jung},
\newblock \bibinfo{title}{Noise-aware evaluation of object detectors},
\newblock in: \bibinfo{booktitle}{Proc. IEEE Winter Conf. Appl. Comput. Vis.
  (WACV)}, \bibinfo{address}{Tucson, AZ, USA}, \bibinfo{year}{2025}, pp.
  \bibinfo{pages}{9322--9331}.
\bibitem[{Su et~al.(2012)Su, Deng, and Fei-Fei}]{crowdsourcing2012su}
\bibinfo{author}{H.~Su}, \bibinfo{author}{J.~Deng},
  \bibinfo{author}{L.~Fei-Fei},
\newblock \bibinfo{title}{Crowdsourcing annotations for visual object
  detection},
\newblock in: \bibinfo{booktitle}{AAAI Workshop, Human Computation},
  \bibinfo{address}{Ontario, Canada}, \bibinfo{year}{2012}, pp.
  \bibinfo{pages}{40--46}.
\bibitem[{Guo et~al.(2017)Guo, Pleiss, Sun, and Weinberger}]{guo2017oncalib}
\bibinfo{author}{C.~Guo}, \bibinfo{author}{G.~Pleiss},
  \bibinfo{author}{Y.~Sun}, \bibinfo{author}{K.~Q. Weinberger},
\newblock \bibinfo{title}{On calibration of modern neural networks},
\newblock in: \bibinfo{booktitle}{Proc. Int. Conf. Mach. Learn. (ICML)},
  \bibinfo{address}{Sydney, Australia}, \bibinfo{year}{2017}, pp.
  \bibinfo{pages}{1321--1330}.
\bibitem[{Kumar et~al.(2019)Kumar, Liang, and Ma}]{kumar2019verif}
\bibinfo{author}{A.~Kumar}, \bibinfo{author}{P.~S. Liang},
  \bibinfo{author}{T.~Ma},
\newblock \bibinfo{title}{Verified uncertainty calibration},
\newblock in: \bibinfo{booktitle}{Proc. Adv. Neural Inform. Process. Syst.
  (NeurIPS)}, volume~\bibinfo{volume}{32}, \bibinfo{address}{Vancouver,
  Canada}, \bibinfo{year}{2019}.
\bibitem[{Nixon et~al.(2019)Nixon, Dusenberry, Zhang, Jerfel, and
  Tran}]{nixon2019measur}
\bibinfo{author}{J.~Nixon}, \bibinfo{author}{M.~W. Dusenberry},
  \bibinfo{author}{L.~Zhang}, \bibinfo{author}{G.~Jerfel},
  \bibinfo{author}{D.~Tran},
\newblock \bibinfo{title}{Measuring calibration in deep learning},
\newblock in: \bibinfo{booktitle}{IEEE Conf. Comput. Vis. Pattern Recog. (CVPR)
  Worksh.}, \bibinfo{address}{Long Beach, CA, USA}, \bibinfo{year}{2019}, pp.
  \bibinfo{pages}{38--41}.
\bibitem[{Mukhoti et~al.(2020)Mukhoti, Kulharia, Sanyal, Golodetz, Torr, and
  Dokania}]{mukh2020calib}
\bibinfo{author}{J.~Mukhoti}, \bibinfo{author}{V.~Kulharia},
  \bibinfo{author}{A.~Sanyal}, \bibinfo{author}{S.~Golodetz},
  \bibinfo{author}{P.~H.~S. Torr}, \bibinfo{author}{P.~K. Dokania},
\newblock \bibinfo{title}{Calibrating deep neural networks using focal loss},
\newblock in: \bibinfo{booktitle}{Proc. Adv. Neural Inform. Process. Syst.
  (NeurIPS)}, \bibinfo{year}{2020}, pp. \bibinfo{pages}{15288--15299}.
\bibitem[{Wang et~al.(2021)Wang, Feng, and Zhang}]{wang2021rethink}
\bibinfo{author}{D.-B. Wang}, \bibinfo{author}{L.~Feng}, \bibinfo{author}{M.-L.
  Zhang},
\newblock \bibinfo{title}{Rethinking calibration of deep neural networks: Do
  not be afraid of overconfidence},
\newblock in: \bibinfo{booktitle}{Proc. Adv. Neural Inform. Process. Syst.
  (NeurIPS)}, \bibinfo{year}{2021}, pp. \bibinfo{pages}{11809--11820}.
\bibitem[{Kuzucu et~al.(2024)Kuzucu, Oksuz, Sadeghi, and
  Dokania}]{kuzucu2024calibration}
\bibinfo{author}{S.~Kuzucu}, \bibinfo{author}{K.~Oksuz},
  \bibinfo{author}{J.~Sadeghi}, \bibinfo{author}{P.~K. Dokania},
\newblock \bibinfo{title}{On calibration of object detectors: Pitfalls,
  evaluation and baselines},
\newblock in: \bibinfo{booktitle}{Proc. Eur. Conf. Comput. Vis. (ECCV)},
  \bibinfo{address}{Milan, Italy}, \bibinfo{year}{2024}.
\bibitem[{Park et~al.(2026)Park, Sobolewski, and Azizan}]{park2026uncertainty}
\bibinfo{author}{Y.-J. Park}, \bibinfo{author}{C.~Sobolewski},
  \bibinfo{author}{N.~Azizan}, \bibinfo{title}{Uncertainty quantification in
  detection transformers: Object-level calibration and image-level
  reliability}, \bibinfo{year}{2026}.
  \href{http://arxiv.org/abs/2412.01782}{{\tt arXiv:2412.01782}}.
\bibitem[{Munir et~al.(2022)Munir, Khan, Sarfraz, and Ali}]{munir2022towards}
\bibinfo{author}{M.~A. Munir}, \bibinfo{author}{M.~H. Khan},
  \bibinfo{author}{M.~Sarfraz}, \bibinfo{author}{M.~Ali},
\newblock \bibinfo{title}{Towards improving calibration in object detection
  under domain shift},
\newblock in: \bibinfo{booktitle}{Proc. Adv. Neural Inform. Process. Syst.
  (NeurIPS)}, \bibinfo{address}{New Orleans, LA, USA}, \bibinfo{year}{2022},
  pp. \bibinfo{pages}{38706--38718}.
\bibitem[{Pathiraja et~al.(2023)Pathiraja, Gunawardhana, and
  Khan}]{pathi2023multiclass}
\bibinfo{author}{B.~Pathiraja}, \bibinfo{author}{M.~Gunawardhana},
  \bibinfo{author}{M.~H. Khan},
\newblock \bibinfo{title}{Multiclass confidence and localization calibration
  for object detection},
\newblock in: \bibinfo{booktitle}{Proc. IEEE Conf. Comput. Vis. Pattern Recog.
  (CVPR)}, \bibinfo{address}{Vancouver, Canada}, \bibinfo{year}{2023}, pp.
  \bibinfo{pages}{19734--19743}.
\bibitem[{Munir et~al.(2023{\natexlab{a}})Munir, Khan, Khan, Ali, and
  Khan}]{munir2023caldetr}
\bibinfo{author}{M.~A. Munir}, \bibinfo{author}{S.~Khan},
  \bibinfo{author}{M.~H. Khan}, \bibinfo{author}{M.~Ali},
  \bibinfo{author}{F.~Khan},
\newblock \bibinfo{title}{Cal-{DETR}: Calibrated detection transformer},
\newblock in: \bibinfo{booktitle}{Proc. Adv. Neural Inform. Process. Syst.
  (NeurIPS)}, \bibinfo{address}{New Orleans, LA, USA},
  \bibinfo{year}{2023}{\natexlab{a}}.
\bibitem[{Munir et~al.(2023{\natexlab{b}})Munir, Khan, Khan, and
  Khan}]{munir2023bridg}
\bibinfo{author}{M.~A. Munir}, \bibinfo{author}{M.~H. Khan},
  \bibinfo{author}{S.~Khan}, \bibinfo{author}{F.~S. Khan},
\newblock \bibinfo{title}{Bridging precision and confidence: A train-time loss
  for calibrating object detection},
\newblock in: \bibinfo{booktitle}{Proc. IEEE Conf. Comput. Vis. Pattern Recog.
  (CVPR)}, \bibinfo{address}{Vancouver, Canada},
  \bibinfo{year}{2023}{\natexlab{b}}, pp. \bibinfo{pages}{11474--11483}.
\bibitem[{Popordanoska et~al.(2024)Popordanoska, Tiulpin, and
  Blaschko}]{popor2024beyondcl}
\bibinfo{author}{T.~Popordanoska}, \bibinfo{author}{A.~Tiulpin},
  \bibinfo{author}{M.~B. Blaschko},
\newblock \bibinfo{title}{Beyond classification: Definition and density-based
  estimation of calibration in object detection},
\newblock in: \bibinfo{booktitle}{Proc. IEEE Winter Conf. Appl. Comput. Vis.
  (WACV)}, \bibinfo{address}{Waikoloa, HI, USA}, \bibinfo{year}{2024}, pp.
  \bibinfo{pages}{574--583}.
\bibitem[{Alexandridis et~al.(2025)Alexandridis, Elezi, Deng, Nguyen, and
  Luo}]{alexandridis2024fractal}
\bibinfo{author}{K.~P. Alexandridis}, \bibinfo{author}{I.~Elezi},
  \bibinfo{author}{J.~Deng}, \bibinfo{author}{A.~Nguyen},
  \bibinfo{author}{S.~Luo},
\newblock \bibinfo{title}{Fractal calibration for long-tailed object
  detection},
\newblock in: \bibinfo{booktitle}{Proc. IEEE Conf. Comput. Vis. Pattern Recog.
  (CVPR)}, \bibinfo{address}{Nashville, TN, USA}, \bibinfo{year}{2025}.
\bibitem[{Hall et~al.(2020)Hall, Dayoub, Skinner, Zhang, Miller, Corke,
  Carneiro, Angelova, and Sünderhauf}]{poddef2020hall}
\bibinfo{author}{D.~Hall}, \bibinfo{author}{F.~Dayoub},
  \bibinfo{author}{J.~Skinner}, \bibinfo{author}{H.~Zhang},
  \bibinfo{author}{D.~Miller}, \bibinfo{author}{P.~Corke},
  \bibinfo{author}{G.~Carneiro}, \bibinfo{author}{A.~Angelova},
  \bibinfo{author}{N.~Sünderhauf},
\newblock \bibinfo{title}{Probabilistic object detection: Definition and
  evaluation},
\newblock in: \bibinfo{booktitle}{Proc. IEEE Winter Conf. Appl. Comput. Vis.
  (WACV)}, \bibinfo{address}{Snowmass Village, CO, USA}, \bibinfo{year}{2020},
  pp. \bibinfo{pages}{1020--1029}.
\bibitem[{Feng et~al.(2022)Feng, Harakeh, Waslander, and
  Dietmayer}]{probobjdetreview2022}
\bibinfo{author}{D.~Feng}, \bibinfo{author}{A.~Harakeh}, \bibinfo{author}{S.~L.
  Waslander}, \bibinfo{author}{K.~Dietmayer},
\newblock \bibinfo{title}{A review and comparative study on probabilistic
  object detection in autonomous driving},
\newblock \bibinfo{journal}{IEEE Trans Intell. Transportation Syst.}
  \bibinfo{volume}{23} (\bibinfo{year}{2022}) \bibinfo{pages}{9961--9980}.
\bibitem[{Miller et~al.(2018)Miller, Nicholson, Dayoub, and
  Sünderhauf}]{dropoutsamp2018}
\bibinfo{author}{D.~Miller}, \bibinfo{author}{L.~Nicholson},
  \bibinfo{author}{F.~Dayoub}, \bibinfo{author}{N.~Sünderhauf},
\newblock \bibinfo{title}{Dropout sampling for robust object detection in
  open-set conditions},
\newblock in: \bibinfo{booktitle}{Proc. IEEE Int. Conf. Robot. Automat.
  (ICRA)}, \bibinfo{address}{Brisbane, Australia}, \bibinfo{year}{2018}, pp.
  \bibinfo{pages}{3243--3249}.
\bibitem[{Miller et~al.(2019)Miller, Dayoub, Milford, and
  Sünderhauf}]{evalmergestrat2019}
\bibinfo{author}{D.~Miller}, \bibinfo{author}{F.~Dayoub},
  \bibinfo{author}{M.~Milford}, \bibinfo{author}{N.~Sünderhauf},
\newblock \bibinfo{title}{Evaluating merging strategies for sampling-based
  uncertainty techniques in object detection},
\newblock in: \bibinfo{booktitle}{Proc. IEEE Int. Conf. Robot. Automat.
  (ICRA)}, \bibinfo{address}{Montreal, Canada}, \bibinfo{year}{2019}, pp.
  \bibinfo{pages}{2348--2354}.
\bibitem[{Lyu et~al.(2020)Lyu, Gutierrez, Rajguru, and
  Beksi}]{probdeepensemble2020}
\bibinfo{author}{Z.~Lyu}, \bibinfo{author}{N.~Gutierrez},
  \bibinfo{author}{A.~Rajguru}, \bibinfo{author}{W.~J. Beksi},
\newblock \bibinfo{title}{Probabilistic object detection via deep ensembles},
\newblock in: \bibinfo{booktitle}{Eur. Conf. Comput. Vis. (ECCV) Workshop},
  \bibinfo{year}{2020}, pp. \bibinfo{pages}{67--75}.
\bibitem[{Neal(1996)}]{bayesneurnetwork}
\bibinfo{author}{R.~M. Neal}, \bibinfo{title}{Bayesian Learning for Neural
  Networks}, \bibinfo{publisher}{Springer-Verlag}, \bibinfo{year}{1996}.
\bibitem[{Le et~al.(2018)Le, Diehl, Brunner, and Knoll}]{uncestdeepobjdet2018}
\bibinfo{author}{M.~T. Le}, \bibinfo{author}{F.~Diehl},
  \bibinfo{author}{T.~Brunner}, \bibinfo{author}{A.~Knoll},
\newblock \bibinfo{title}{Uncertainty estimation for deep neural object
  detectors in safety-critical applications},
\newblock in: \bibinfo{booktitle}{Proc. IEEE Intell. Transportation Syst.
  Conf.}, \bibinfo{address}{Maui, HA, USA}, \bibinfo{year}{2018}, pp.
  \bibinfo{pages}{3873--3878}.
\bibitem[{Choi et~al.(2019)Choi, Chun, Kim, and Lee}]{gaussyolov32019}
\bibinfo{author}{J.~Choi}, \bibinfo{author}{D.~Chun}, \bibinfo{author}{H.~Kim},
  \bibinfo{author}{H.-J. Lee},
\newblock \bibinfo{title}{Gaussian {YOLOv3}: An accurate and fast object
  detector using localization uncertainty for autonomous driving},
\newblock in: \bibinfo{booktitle}{Proc. IEEE Int. Conf. Comput. Vis. (ICCV)},
  \bibinfo{address}{Seoul, Korea}, \bibinfo{year}{2019}, pp.
  \bibinfo{pages}{502--511}.
\bibitem[{He et~al.(2019)He, Zhu, Wang, Savvides, and Zhang}]{bboxregunc2019he}
\bibinfo{author}{Y.~He}, \bibinfo{author}{C.~Zhu}, \bibinfo{author}{J.~Wang},
  \bibinfo{author}{M.~Savvides}, \bibinfo{author}{X.~Zhang},
\newblock \bibinfo{title}{Bounding box regression with uncertainty for accurate
  object detection},
\newblock in: \bibinfo{booktitle}{Proc. IEEE Conf. Comput. Vis. Pattern Recog.
  (CVPR)}, \bibinfo{address}{Long Beach, CA, USA}, \bibinfo{year}{2019}, pp.
  \bibinfo{pages}{2883--2892}.
\bibitem[{Kraus and Dietmayer(2019)}]{unconestage2019}
\bibinfo{author}{F.~Kraus}, \bibinfo{author}{K.~Dietmayer},
\newblock \bibinfo{title}{Uncertainty estimation in one-stage object
  detection},
\newblock in: \bibinfo{booktitle}{Proc. IEEE Intell. Transportation Syst.
  Conf.}, \bibinfo{address}{Auckland, New Zealand}, \bibinfo{year}{2019}, pp.
  \bibinfo{pages}{53--60}.
\bibitem[{Harakeh et~al.(2020)Harakeh, Smart, and Waslander}]{bayesod2020}
\bibinfo{author}{A.~Harakeh}, \bibinfo{author}{M.~Smart},
  \bibinfo{author}{S.~L. Waslander},
\newblock \bibinfo{title}{Bayesod: A bayesian approach for uncertainty
  estimation in deep object detectors},
\newblock in: \bibinfo{booktitle}{Proc. IEEE Int. Conf. Robot. Automat.
  (ICRA)}, \bibinfo{year}{2020}, pp. \bibinfo{pages}{87--93}.
\bibitem[{He and Wang(2020)}]{deepmixden2020}
\bibinfo{author}{Y.~He}, \bibinfo{author}{J.~Wang},
\newblock \bibinfo{title}{Deep mixture density network for probabilistic object
  detection},
\newblock in: \bibinfo{booktitle}{Proc. IEEE/RSJ Int. Conf. Intell. Robots
  Syst. (IROS)}, \bibinfo{year}{2020}, pp. \bibinfo{pages}{10550--10555}.
\bibitem[{Harakeh and Waslander(2021)}]{estandevaldod2021}
\bibinfo{author}{A.~Harakeh}, \bibinfo{author}{S.~L. Waslander},
\newblock \bibinfo{title}{Estimating and evaluating regression predictive
  uncertainty in deep object detectors},
\newblock in: \bibinfo{booktitle}{Proc. Int. Conf. Learn. Represent. (ICLR)},
  \bibinfo{year}{2021}.
\bibitem[{Su et~al.(2023)Su, Li, He, Han, Feng, Ding, and
  Miao}]{uncquanselfdrive2023su}
\bibinfo{author}{S.~Su}, \bibinfo{author}{Y.~Li}, \bibinfo{author}{S.~He},
  \bibinfo{author}{S.~Han}, \bibinfo{author}{C.~Feng},
  \bibinfo{author}{C.~Ding}, \bibinfo{author}{F.~Miao},
\newblock \bibinfo{title}{Uncertainty quantification of collaborative detection
  for self-driving},
\newblock in: \bibinfo{booktitle}{Proc. IEEE Int. Conf. Robot. Automat.
  (ICRA)}, \bibinfo{address}{London, United Kingdom}, \bibinfo{year}{2023}, pp.
  \bibinfo{pages}{5588--5594}.
\bibitem[{Ausiello et~al.(2013)Ausiello, Crescenzi, Gambosi, Kann,
  Marchetti-Spaccamela, and Protasi}]{apx}
\bibinfo{author}{G.~Ausiello}, \bibinfo{author}{P.~Crescenzi},
  \bibinfo{author}{G.~Gambosi}, \bibinfo{author}{V.~Kann},
  \bibinfo{author}{A.~Marchetti-Spaccamela}, \bibinfo{author}{M.~Protasi},
  \bibinfo{title}{Complexity and Approximation: Combinatorial Optimization
  Problems and Their Approximability Properties}, \bibinfo{publisher}{Springer
  Publishing Company}, \bibinfo{year}{2013}.
\bibitem[{Mahalanobis(1936)}]{mahalanobis1936generalized}
\bibinfo{author}{P.~C. Mahalanobis},
\newblock \bibinfo{title}{On the generalized distance in statistics},
\newblock \bibinfo{journal}{Proceedings of the National Institute of Sciences
  of India} \bibinfo{volume}{2} (\bibinfo{year}{1936}) \bibinfo{pages}{49--55}.
\bibitem[{Tsybakov(2009)}]{tvd}
\bibinfo{author}{A.~B. Tsybakov}, \bibinfo{title}{Introduction to Nonparametric
  Estimation}, Springer Series in Statistics, \bibinfo{publisher}{Springer},
  \bibinfo{address}{New York, NY, USA}, \bibinfo{year}{2009},
  p.~\bibinfo{pages}{83}.
\bibitem[{Krishnamoorthy(2006)}]{conf_interval}
\bibinfo{author}{K.~Krishnamoorthy},
\newblock \bibinfo{title}{Normal distribution},
\newblock in: \bibinfo{booktitle}{Handbook of Statistical Distributions with
  Applications}, \bibinfo{publisher}{CRC Press}, \bibinfo{address}{Boca Raton,
  FL, USA}, \bibinfo{year}{2006}, p. \bibinfo{pages}{151}.
\bibitem[{Ren et~al.(2015)Ren, He, Girshick, and Sun}]{fasterrcnn2015}
\bibinfo{author}{S.~Ren}, \bibinfo{author}{K.~He},
  \bibinfo{author}{R.~Girshick}, \bibinfo{author}{J.~Sun},
\newblock \bibinfo{title}{Faster {R-CNN}: Towards real-time object detection
  with region proposal networks},
\newblock in: \bibinfo{booktitle}{Proc. Adv. Neural Inform. Process. Syst.
  (NeurIPS)}, \bibinfo{address}{Montr\'{e}al, Canada}, \bibinfo{year}{2015},
  pp. \bibinfo{pages}{91--99}.
\bibitem[{Ge et~al.(2021)Ge, Liu, Wang, Li, and Sun}]{yolox2021}
\bibinfo{author}{Z.~Ge}, \bibinfo{author}{S.~Liu}, \bibinfo{author}{F.~Wang},
  \bibinfo{author}{Z.~Li}, \bibinfo{author}{J.~Sun}, \bibinfo{title}{{YOLOX}:
  Exceeding {YOLO} series in 2021}, \bibinfo{year}{2021}.
  \href{http://arxiv.org/abs/2107.08430}{{\tt arXiv:2107.08430}}.
\bibitem[{Wang et~al.(2023)Wang, Bochkovskiy, and Liao}]{wang2022yolov7}
\bibinfo{author}{C.-Y. Wang}, \bibinfo{author}{A.~Bochkovskiy},
  \bibinfo{author}{H.-Y.~M. Liao},
\newblock \bibinfo{title}{{YOLOv7}: Trainable bag-of-freebies sets new
  state-of-the-art for real-time object detectors},
\newblock in: \bibinfo{booktitle}{Proc. IEEE Conf. Comput. Vis. Pattern Recog.
  (CVPR)}, \bibinfo{address}{Vancouver, Canada}, \bibinfo{year}{2023}, pp.
  \bibinfo{pages}{7464--7475}.
\bibitem[{Lin et~al.(2017)Lin, Goyal, Girshick, He, and Dollar}]{retinanet2017}
\bibinfo{author}{T.-Y. Lin}, \bibinfo{author}{P.~Goyal},
  \bibinfo{author}{R.~Girshick}, \bibinfo{author}{K.~He},
  \bibinfo{author}{P.~Dollar},
\newblock \bibinfo{title}{Focal loss for dense object detection},
\newblock in: \bibinfo{booktitle}{Proc. IEEE Int. Conf. Comput. Vis. (ICCV)},
  \bibinfo{address}{Venice, Italy}, \bibinfo{year}{2017}.
\bibitem[{Robertson et~al.(1988)Robertson, Wright, and
  Dykstra}]{robert1988isotonicreg}
\bibinfo{author}{T.~Robertson}, \bibinfo{author}{F.~T. Wright},
  \bibinfo{author}{R.~Dykstra}, \bibinfo{title}{Order Restricted Statistical
  Inference}, \bibinfo{publisher}{Wiley}, \bibinfo{year}{1988}.
\bibitem[{Pedregosa et~al.(2011)Pedregosa, Varoquaux, Gramfort, Michel,
  Thirion, Grisel, Blondel, Prettenhofer, Weiss, Dubourg, Vanderplas, Passos,
  Cournapeau, Brucher, Perrot, and Duchesnay}]{scikitlearn}
\bibinfo{author}{F.~Pedregosa}, \bibinfo{author}{G.~Varoquaux},
  \bibinfo{author}{A.~Gramfort}, \bibinfo{author}{V.~Michel},
  \bibinfo{author}{B.~Thirion}, \bibinfo{author}{O.~Grisel},
  \bibinfo{author}{M.~Blondel}, \bibinfo{author}{P.~Prettenhofer},
  \bibinfo{author}{R.~Weiss}, \bibinfo{author}{V.~Dubourg},
  \bibinfo{author}{J.~Vanderplas}, \bibinfo{author}{A.~Passos},
  \bibinfo{author}{D.~Cournapeau}, \bibinfo{author}{M.~Brucher},
  \bibinfo{author}{M.~Perrot}, \bibinfo{author}{E.~Duchesnay},
\newblock \bibinfo{title}{Scikit-learn: Machine learning in {P}ython},
\newblock \bibinfo{journal}{J. Mach. Learn. Res. (JMLR)} \bibinfo{volume}{12}
  (\bibinfo{year}{2011}) \bibinfo{pages}{2825--2830}.
\bibitem[{Kendall and Gal(2017)}]{kendall_prob2017}
\bibinfo{author}{A.~Kendall}, \bibinfo{author}{Y.~Gal},
\newblock \bibinfo{title}{What uncertainties do we need in bayesian deep
  learning for computer vision?},
\newblock in: \bibinfo{booktitle}{Proc. Adv. Neural Inform. Process. Syst.
  (NeurIPS)}, \bibinfo{address}{Long Beach, California, USA},
  \bibinfo{year}{2017}, pp. \bibinfo{pages}{5580--–5590}.
\bibitem[{Paszke et~al.(2019)Paszke, Gross, Massa, Lerer, Bradbury, Chanan,
  Killeen, Lin, Gimelshein, Antiga, Desmaison, Kopf, Yang, DeVito, Raison,
  Tejani, Chilamkurthy, Steiner, Fang, Bai, and Chintala}]{pytorch}
\bibinfo{author}{A.~Paszke}, \bibinfo{author}{S.~Gross},
  \bibinfo{author}{F.~Massa}, \bibinfo{author}{A.~Lerer},
  \bibinfo{author}{J.~Bradbury}, \bibinfo{author}{G.~Chanan},
  \bibinfo{author}{T.~Killeen}, \bibinfo{author}{Z.~Lin},
  \bibinfo{author}{N.~Gimelshein}, \bibinfo{author}{L.~Antiga},
  \bibinfo{author}{A.~Desmaison}, \bibinfo{author}{A.~Kopf},
  \bibinfo{author}{E.~Yang}, \bibinfo{author}{Z.~DeVito},
  \bibinfo{author}{M.~Raison}, \bibinfo{author}{A.~Tejani},
  \bibinfo{author}{S.~Chilamkurthy}, \bibinfo{author}{B.~Steiner},
  \bibinfo{author}{L.~Fang}, \bibinfo{author}{J.~Bai},
  \bibinfo{author}{S.~Chintala},
\newblock \bibinfo{title}{Pytorch: An imperative style, high-performance deep
  learning library},
\newblock in: \bibinfo{booktitle}{Proc. Adv. Neural Inform. Process. Syst.
  (NeurIPS)}, \bibinfo{address}{Vancouver, Canada}, \bibinfo{year}{2019}, pp.
  \bibinfo{pages}{8024--8035}.
\bibitem[{Lakshminarayanan et~al.(2017)Lakshminarayanan, Pritzel, and
  Blundell}]{deep_ensemble}
\bibinfo{author}{B.~Lakshminarayanan}, \bibinfo{author}{A.~Pritzel},
  \bibinfo{author}{C.~Blundell},
\newblock \bibinfo{title}{Simple and scalable predictive uncertainty estimation
  using deep ensembles},
\newblock in: \bibinfo{booktitle}{Proc. Adv. Neural Inform. Process. Syst.
  (NeurIPS)}, \bibinfo{address}{Long Beach, California, USA},
  \bibinfo{year}{2017}, pp. \bibinfo{pages}{6405–--6416}.
\bibitem[{Oksuz et~al.(2022)Oksuz, Cam, Kalkan, and Akbas}]{lrp}
\bibinfo{author}{K.~Oksuz}, \bibinfo{author}{B.~C. Cam},
  \bibinfo{author}{S.~Kalkan}, \bibinfo{author}{E.~Akbas},
\newblock \bibinfo{title}{One metric to measure them all: Localisation recall
  precision ({LRP}) for evaluating visual detection tasks},
\newblock \bibinfo{journal}{IEEE Trans. Pattern Anal. Mach. Intell. (TPAMI)}
  \bibinfo{volume}{44} (\bibinfo{year}{2022}) \bibinfo{pages}{9446--9463}.

\end{thebibliography}
}

\end{document}